\documentclass[5p]{elsarticle}

\usepackage[dvipdfmx]{}
\usepackage{hyperref}
\usepackage{bm}
\usepackage{amsmath}
\usepackage{amssymb}
\usepackage{algorithm}
\usepackage{algcompatible}
\usepackage{array}
\usepackage{eqparbox}
\usepackage{siunitx}
\usepackage{wrapfig}
\usepackage{times}
\usepackage{url}
\usepackage{color}
\newcommand{\switchlanguage}[2]{%
  \ifx\paperlanguage\empty%
  #1%
  \else%
  #2%
  \fi%
}
\newcommand{\secref}[1]{Sec. \ref{#1}}
\newcommand{\tabref}[1]{{Table \ref{#1}}}
\newcommand{\figref}[1]{{Fig. \ref{#1}}}
\newcommand{\equref}[1]{{Eq. \ref{#1}}}

\def\paperlanguage{}

\journal{Robotics and Autonomous Systems}

\begin{document}

\begin{frontmatter}

\title{
  \switchlanguage%
  {%
    Robust Continuous Motion Strategy Against Muscle Rupture using Online Learning of Redundant Intersensory Networks for Musculoskeletal Humanoids
  }%
  {%
    筋骨格ヒューマノイドの冗長性なセンサ間ネットワークのオンライン学習を用いた筋破断に対するロバストな継続的動作戦略
  }%
}

\author{Kento Kawaharazuka\corref{author}\fnref{jskfootnote}}
\ead{kawaharazuka@jsk.imi.i.u-tokyo.ac.jp}
\author{Manabu Nishiura\fnref{jskfootnote}}
\author{Yasunori Toshimitsu\fnref{jskfootnote}}
\author{Yusuke Omura\fnref{jskfootnote}}
\author{\\Yuya Koga\fnref{jskfootnote}}
\author{Yuki Asano\fnref{jskfootnote}}
\author{Kei Okada\fnref{jskfootnote}}
\author{Koji Kawasaki\fnref{toyotafootnote}}
\author{Masayuki Inaba\fnref{jskfootnote}}
\address[jskfootnote]{Department of Mechano-Informatics, Graduate School of Information Science and Technology, The University of Tokyo, Japan}
\address[toyotafootnote]{TOYOTA Motor Corporation, Aichi 471-8751, Japan}
\cortext[author]{Corresponding author.}

\begin{abstract}
\switchlanguage%
{%
  Musculoskeletal humanoids have various biomimetic advantages, of which redundant muscle arrangement is one of the most important features.
  This feature enables variable stiffness control and allows the robot to keep moving its joints even if one of the redundant muscles breaks, but this has been rarely explored.
  In this study, we construct a neural network that represents the relationship among sensors in the flexible and difficult-to-modelize body of the musculoskeletal humanoid, and by learning this neural network, accurate motions can be achieved.
  In order to take advantage of the redundancy of muscles, we discuss the use of this network for muscle rupture detection, online update of the intersensory relationship considering the muscle rupture, and body control and state estimation using the muscle rupture information.
  This study explains a method of constructing a musculoskeletal humanoid that continues to move and perform tasks robustly even when one muscle breaks.
}%
{%
  筋骨格ヒューマノイドは様々な生物規範型の利点を有するが, その中でも冗長な筋配置は最も重要な特徴の一つである.
  この特徴は可変剛性制御を可能にすると同時に, 冗長な筋が一本切れてもロボットが関節を動かし続けることを可能とするが, この利点はほとんど探索されていない.
  本研究では, この筋骨格ヒューマノイドの柔軟でモデル化困難な身体に存在するセンサ群の関係性を表現するニューラルネットワークを構築し, これを学習することで正確な動作を実現する.
  また, この枠組みの中で筋の冗長性の利点を最大限に活かすため, このネットワークを使った筋の異常検知, 筋破断時におけるセンサ間関係の再学習, 筋破断時における身体制御・状態推定等について議論する.
  本研究により, 筋が一本切れても身体を動かし続け, よりロバストにタスクを継続する筋骨格ヒューマノイドの構成法を明らかにした.
}%
\end{abstract}


\end{frontmatter}


\section{Introduction}\label{sec:introduction}
\begin{figure}[t]
  \centering
  \includegraphics[width=0.99\columnwidth]{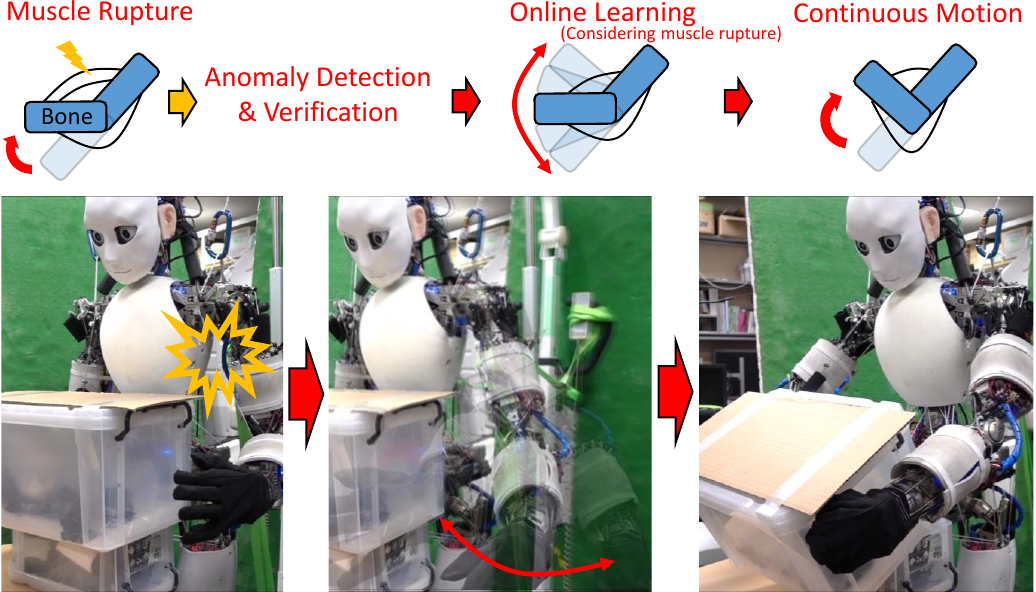}
  \caption{The overall concept of this study: anomaly detection for muscle rupture, verification of the muscle rupture state, online learning of intersensory networks considering muscle rupture, and robust continuous motion control using them.}
  \label{figure:motivation}
\end{figure}

\switchlanguage%
{%
  The musculoskeletal humanoid \cite{kawaharazuka2019musashi, nakanishi2013design, wittmeier2013toward, jantsch2013anthrob, asano2016kengoro} has various biomimetic advantages such as variable stiffness using redundant muscles, ball joints without singular points, underactuated and flexible fingers, etc.
  In this study, we develop a robust learning control strategy focusing on the muscle redundancy, which is one of the characteristics of musculoskeletal humanoids.
  In a musculoskeletal structure where the control input (muscle length) is redundant with respect to the joint angle to be controlled, even if one muscle is broken, it can be compensated and the robot can continue to move as before, leading to a more robust robot configuration (\figref{figure:motivation}).
  In order to make use of this characteristic for musculoskeletal humanoids, which are flexible and difficult to modelize, we propose a new learning-based framework.
  In this study, online learning of a musculoskeletal intersensory network, as well as muscle length-based control, state estimation, and anomaly detection using it are described, forming a strategy against muscle rupture.
  The muscle redundancy is utilized to realize continuous motions in an actual musculoskeletal humanoid, Musashi \cite{kawaharazuka2019musashi}.
  From the experiments, we show that the algorithm of online learning, state estimation, and control can be modified based on the information of muscle rupture to adapt to the current body state, without destroying the network structure.
  This enables a series of continuous movements even if one muscle is broken.
}%
{%
  筋骨格ヒューマノイド\cite{kawaharazuka2019musashi, nakanishi2013design, wittmeier2013toward, jantsch2013anthrob, asano2016kengoro}は冗長な筋肉による可変剛性・特異点のない球関節・劣駆動で柔軟な指等様々な生物模倣型の利点を有する.
  その中でも本研究では, 筋骨格ヒューマノイドの特性である, 筋の冗長性に着目した制御戦略について述べる.
  制御対象である関節角度に対して制御入力(筋長)が冗長である筋骨格構造においては, 筋が一本切れたとしても, それを補償してこれまでと同じように動作し続けることが可能であり, よりロバストなロボット構成へと繋がる(\figref{figure:motivation}).
  この戦略を, 柔軟でモデル化が困難な筋骨格ヒューマノイドに適用するため, 本研究では学習型制御を用いた新しい枠組みを提案する.
  筋破断検知・筋長制御・状態推定とそのオンライン学習, 筋破断時の戦略について述べ, この筋冗長性を活かした動作を筋骨格ヒューマノイドMusashiの実機において実現する.
  実験により, 筋破断時には, オンライン学習・状態推定・筋長制御のアルゴリズムを筋破断情報を元に変更することで, 現状のネットワーク構造を壊さずに現在の身体状態に適応可能であること, これにより途中で筋が一本切れても一連の継続的動作が可能となることを示した.
}%

\subsection{Redundancy of Musculoskeletal Structures}\label{subsec:redundancy}
\switchlanguage%
{%
  The muscle redundancy has been mainly focused on variable stiffness control \cite{kawaharazuka2019longtime} and biarticular muscles \cite{sharbafi2016biarticular} so far.
  Also, since the muscle redundancy is difficult to handle for control in some cases, many methods have been proposed to eliminate it \cite{zhong2019reducing, allessandro2013synergy}.
  On the other hand, another important feature of the muscle redundancy is that the robot can continue to move even if one of the muscles is broken, but this has been rarely explored.
  \cite{kawamura2016jointspace} refers to a possibility to include information of muscle rupture in muscle tension-based control, but there is no evaluation of this feature, no discussion on judging whether the muscle is broken or not (anomaly detection and its verification), and no relearning of the body model after the muscle rupture.
}%
{%
  これまで筋の冗長性は主に, 可変剛性\cite{kawaharazuka2019longtime}や二関節筋\cite{sharbafi2016biarticular}という点に対して焦点が当てられてきた.
  また, 大抵の場合は冗長性は制御にとって邪魔であるため, 冗長性を排除する方法が多く提案されている\cite{zhong2019reducing, allessandro2013synergy}.
  一方, その他の筋の冗長性の特徴として, 筋が一本切れても動く, という重要な特徴があるが, これはほとんど探索されていない.
  \cite{kawamura2016jointspace}では筋張力制御において, 筋が切れているかどうかの情報を含めることが可能とあるが, それらについての評価はなく, 筋が切れているかどうかの判断(異常検知)や破断後の身体変化の再学習等に関する議論もない.
}%

\subsection{Control, State Estimation, and Anomaly Detection for Flexible Musculoskeletal Structures}\label{subsec:controller}
\switchlanguage%
{%
  Various studies have been conducted to control the flexible musculoskeletal structure which is difficult to modelize.
  There are some studies \cite{jantsch2011scalable, jantsch2012computed, kawamura2016jointspace} of muscle tension-based control methods.
  These studies make use of joint-muscle mappings \cite{ookubo2015learning} learned in actual robots as muscle Jacobian.
  However, it is difficult to control such robots as intended due to the large friction and hysteresis between the muscles and various parts of the body \cite{kawamura2016jointspace}.
  Therefore, this study is based on muscle length-based control.

  As a method of muscle length-based control, Motegi et al. proposed a method to realize an accurate hand position by updating a table that correlates the position with muscle length \cite{motegi2012jacobian}.
  Mizuuchi et al. represented the relationship between joints and muscles with neural networks from motion capture data and performed accurate body control \cite{mizuuchi2006acquisition}.
  Kawaharazuka et al. have developed a method to control the body by learning a neural network that represents the relationship between joints and muscles online \cite{kawaharazuka2018online}, and extended it to take into account the change in muscle route due to the flexibility of the body tissue \cite{kawaharazuka2019longtime, kawaharazuka2018bodyimage}.
  This online learning mechanism is essential for musculoskeletal humanoids, where the relationship among body sensors are constantly changing.
  However, \cite{motegi2012jacobian} is unsuitable to consider muscle rupture because it directly rewrites the data table of the hand position and muscle length, and thus all data must be rewritten when one muscle is broken.
  \cite{mizuuchi2006acquisition, kawaharazuka2018online} cannot handle muscle tension.
  \cite{kawaharazuka2019longtime, kawaharazuka2018bodyimage} can handle muscle tension but cannot handle the change in the control command when one muscle is broken, because the muscle length command is obtained by only forwarding a neural network once.

  Other methods have been developed to dynamically control simple pneumatic robots with one or two degrees of freedom using Gaussian Process Regression (GP) \cite{buchler2018gaussian}, neural networks and sequential quadratic programming \cite{driess2018sqp}, etc., but they have only been experimented with low degrees of freedom (even with a robot with many degrees of freedom, different controllers have been created for each joint).
  Also, they have not been able to deal with the physical changes (e.g. muscle rupture) and online learning of the body model (it is especially difficult to deal with online learning and physical changes flexibly in GP).
  If we want to obtain a dynamic model of a complex body such as a musculoskeletal humanoid, we need a large amount of data, and the model is usually overfitted by online learning due to the model complexity.
  For redundant systems other than the musculoskeletal robot, there are options such as dynamic neural networks (DNN) \cite{xie2021cyclic}, dynamic movement primitives (DMP) \cite{huang2016jointly}, and reinforcement learning (RL).
  However, DNN makes a number of assumptions on the body model, DMP is not suitable to relearn the body model online after a change in the body, and RL takes an unrealistically long time to learn for complex actual systems.
  From these perspectives, this study uses muscle length-based control instead of a muscle tension-based one, a static model instead of a dynamic one, and a neural network instead of other modeling methods.

  Musculoskeletal AutoEncoder (MAE) \cite{kawaharazuka2020autoencoder} is used as a core method of this study, which extends \cite{kawaharazuka2019longtime, kawaharazuka2018bodyimage}.
  This allows for a unified method for state estimation, control, and simulation of musculoskeletal humanoids using the mutual relationship among joint angle, muscle tension, and muscle length.
  We utilize the feature that the state estimation and control are performed by optimization based on the minimization of loss function, and we modify the algorithm in order to handle muscle rupture by changing the definition of the loss function.
  For anomaly detection, some methods such as \cite{allessandro2006anomaly, park2018anomaly, principi2019unsupervised} have been proposed, but the classical method \cite{allessandro2006anomaly} requires an accurate dynamic model of the body, and learning models \cite{park2018anomaly, principi2019unsupervised} have been developed as different models from state estimation and control.
  Our method is effective in that it eliminates the need for individual model management, since state estimation, control, and anomaly detection are performed in a single unified network.
  In this study, the functions of MAE have been extended by improving the framework of state estimation, control, and anomaly detection, and by adding the change in all components to make use of the muscle redundancy.
  We call our proposed network Redundant Musculoskeletal AutoEncoder (RMAE).
}%
{%
  これまで, 柔軟でモデリング困難な筋骨格構造を制御するために多くの研究が成されてきた.
  筋張力制御手法として\cite{jantsch2011scalable, jantsch2012computed, kawamura2016jointspace}の研究が存在する.
  これらは, 筋長ヤコビアンとして, 実機により学習された関節-筋空間マッピング\cite{ookubo2015learning}を利用している.
  しかし, 球関節や背骨など多くの部分と筋が接触し摩擦・ヒステリシスが大きいため意図したように制御することは難しく\cite{kawamura2016jointspace}, 本研究では筋長制御を行う.

  筋長制御手法として, 茂木らは, 手先位置と筋長を対応付けるテーブルを構築し, 正確な手先位置を実現する手法を提案している\cite{motegi2012jacobian}.
  また, 水内らは, モーションキャプチャデータから関節と筋の関係をニューラルネットワークで表現し, 身体制御を行っている\cite{mizuuchi2006acquisition}.
  河原塚らは, 関節と筋の関係を表すニューラルネットワークをオンラインで学習し身体制御を行う手法\cite{kawaharazuka2018online}, さらにそれを拡張し, 筋張力の影響による身体組織の柔軟性を考慮した手法を開発している\cite{kawaharazuka2018bodyimage, kawaharazuka2019longtime}.
  常に身体センサ間の相関関係が変化し得る筋骨格ヒューマノイドには, このオンライン学習機構が必須である.
  一方, \cite{motegi2012jacobian}は手先位置と筋長のデータテーブルを直接書き換えるため, 筋が切れた場合には全データを一切利用できなくなってしまうので不向きである.
  \cite{mizuuchi2006acquisition, kawaharazuka2018online}は筋張力を扱うことができず, また制御しか扱うことができない.
  \cite{kawaharazuka2018bodyimage, kawaharazuka2019longtime}は筋張力を扱うことができるものの, 一度のneural networkのforwardingによって筋長指令値を求めるため, 筋が切れた際の制御指令値変化を扱うことができない.

  この他にも, 1,2自由度程度のシンプルな空気圧ロボットをGaussian Process Regression (GP) \cite{buchler2018gaussian}やneural networkとsequential quadratic programming \cite{driess2018sqp}を使って動的に制御する方法が開発されているが, 低い自由度でのみ実験されている(多くの自由度を持つロボットを使っても, 関節ごとに異なるcontrollerを作っている)
  また, 筋破断のような身体変化・オンライン学習等については扱うことができていない(特にGP等のモデルはオンライン学習や身体変化等について柔軟に扱うのが難しい).
  もし筋骨格ヒューマノイドのような複雑な身体の動的モデルを得たい場合, 大量のデータが必要であり, オンライン学習も過学習しやすくなってしまう.
  筋骨格以外の冗長系についてもdynamic neural networkやdynamic movement primitives, reinforcement learning等の選択肢があるものの\cite{xie2021cyclic, huang2016jointly}, それぞれ身体モデルに多くの仮定を置いている, 身体変化後の再学習・オンライン学習が難しい, 複雑な系の実機においては現実的ではないほど学習に時間がかかる等の問題がある.
  これらからの観点から, 本研究では筋張力ではなく筋長制御ベース, また, 動的ではなく静的モデル, そして, オンライン学習に適したneural networkを利用している.

  本研究では, \cite{kawaharazuka2018bodyimage, kawaharazuka2019longtime}の研究をさらに拡張した, Musculoskeletal AutoEncoder \cite{kawaharazuka2020autoencoder}を中核に用いる.
  この手法は, 関節角度・筋張力・筋長の相互関係を用いて, 状態推定, 制御, シミュレーションを統一的に行う方法を提供する.
  損失関数の最小化に基づく最適化によって状態推定や制御を行うため, 損失関数の定義を変化させることで筋破断を扱うことが可能になる.
  また, 静的な身体モデルに限ることで, 動的モデルには難しいオンライン学習を現実的な時間で可能としている.
  異常検知についてはこれまで\cite{allessandro2006anomaly, park2018anomaly, principi2019unsupervised}等の手法が提案されているが, 古典的手法\cite{allessandro2006anomaly}は身体の動的なモデルが必要であり, 学習型の手法は本研究とは大きく異ならないが, 状態推定や制御とは個別にモデルが開発されてきている.
  一方で, 本手法は状態推定・制御・異常検知等を統一的に一つのネットワークで行うため, 個別のモデル管理が必要なくなる点で有効である.
  本研究では, このうち状態推定と制御の枠組みを改良して用い, 筋破断検知の追加, 冗長性を活かした動作戦略へと繋げ, 機能を制限・拡張しているため, Redundant Musculoskeletal AutoEncoder (RMAE)と呼ぶこととする.
}%

\subsection{Our Contribution} \label{subsec:our-contribution}
\switchlanguage%
{%
  This study discusses a new robust motion strategy that explicitly exploits the redundancy of the muscle, which has not been handled much so far.
  The system detects the anomaly, verifies the muscle rupture state, relearns the body model after the muscle rupture, and then resumes the behavior as before.
  A comprehensive discussion is also given on the hardware, software, and network architecture for the robust motion strategy.
  The contributions of this study are as follows.
  \begin{itemize}
    \item Modularized hardware and learning software system design to take advantage of the muscle redundancy
    \item Online learning of the musculoskeletal intersensory network using the mutual relationship among joint angle, muscle tension, and muscle length, as well as control, state estimation, and anomaly detection using the same model
    \item Changes in online learning, control, and state estimation after muscle rupture using muscle rupture verification and changes in loss function with the muscle rupture information
    \item Robust motion experiments using the muscle redundancy of the actual musculoskeletal humanoid
  \end{itemize}

}%
{%
  本研究は, これまで多く扱われてこなかった, 筋の冗長性を陽に利用した新しいロバスト動作戦略について議論する.
  筋が切れたことを検知, 筋破断状態を認識し, 筋破断後のモデルを再学習して, これまでと同じように動作を再開するという一連の動作を行う.
  また, この動作のために必要なハードウェア・ソフトウェアの構成, ネットワーク構成等まで含め, 包括的な議論を行う.
  本研究のcontributionは以下である.
  \begin{itemize}
    \item 冗長性を利用するためのモジュール化されたハードウェア・学習型ソフトウェアの全体システム設計
    \item 筋破断状態の認識と筋破断情報による損失関数の変化を使った筋破断後におけるオンライン学習・制御・状態推定の変化
    \item 筋骨格ヒューマノイドの筋冗長性を利用したロバスト動作実験
  \end{itemize}

  本研究の構成は以下のようになっている.
  \secref{sec:musculoskeletal-structure}では, 筋骨格型ロボットの基本構造について説明する.
  \secref{sec:proposed-methods}では, 本研究の流れ, 筋冗長性を活かすための冗長性/交換性を有するハードウェア構成・学習型ソフトウェア構成について説明する.
  \secref{sec:experiments}では, それぞれのコンポーネントについてシミュレーションと実機において実験し, 一連の流れを両手による物体運搬実験によって示す.
}%

\begin{figure}[t]
  \centering
  \includegraphics[width=0.7\columnwidth]{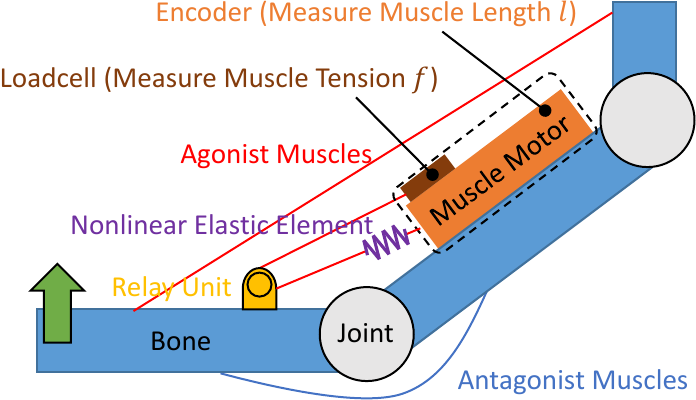}
  \caption{The basic musculoskeletal structure.}
  \label{figure:basic-structure}
\end{figure}

\section{Basic Musculoskeletal Structure} \label{sec:musculoskeletal-structure}
\switchlanguage%
{%
  In the musculoskeletal structure, redundant muscles are antagonistically arranged around the joints (\figref{figure:basic-structure}).
  Not only monoarticular muscles across one joint but also polyarticular muscles across multiple joints exist.
  For the number of degrees of freedom (DOFs) of joints $D$, the minimum number of muscles required to move them is $D+1$, but most musculoskeletal humanoids have more muscles than that.
  In other words, even if one of the muscles is broken, the robot can still move with the other muscles, whereas for axis-driven robots, when one joint is broken, they must stop the movement.
  The muscles are mainly composed of Dyneema, which is an abrasion resistant synthetic fiber.
  Nonlinear elastic elements that allow variable stiffness control are often arranged in series with the muscles.
  In some robots, the muscles are folded back with muscle relay units composed of pulleys in order to gain momentum arm.
  The muscles are sometimes wrapped with soft foam covers for the flexible contact, which make the robot more difficult to modelize.
  For each muscle, an encoder can measure muscle length $l$, a loadcell can measure muscle tension $f$, and a thermal sensor can measure muscle temperature $c$.
  Although joint angles $\bm{\theta}$ cannot be measured in many cases due to the ball joint and complex scapula, some robots can measure them using special mechanisms \cite{kawaharazuka2019musashi, urata2006sensor}.
  Also, by using a vision sensor, it is possible to estimate the joint angle, which can be used as the actual joint angle.
  In this method, AR markers, which can get its six-dimensional position from vision, are attached to the end-effector of the robot.
  The marker position $\bm{P}_{marker}$ is obtained from vision, and the inverse kinematics is solved for $\bm{P}^{ref}=\bm{P}_{marker}$ by using the estimated joint angle $\bm{\theta}^{est}$ from the change in muscle length \cite{ookubo2015learning} as the initial value $\bm{\theta}^{init}$, as shown below,
  \begin{align}
    {\bm{\theta}^{est}}' = \textrm{IK}(\bm{P}^{ref}=\bm{P}_{marker}, \bm{\theta}^{init}=\bm{\theta}^{est})
  \end{align}
  where IK expresses inverse kinematics and ${\bm{\theta}^{est}}'$ is the estimated joint angle corrected by vision.
  Note that this study updates the relationship between joints and muscles, but assumes that the geometric model for joint position and link length is correct.
}%
{%
  筋骨格構造では, 冗長な筋が関節の周りに拮抗して配置されている(\figref{figure:basic-structure}).
  また, 一つの関節を跨ぐ一関節筋だけでなく, 複数の関節を跨ぐ多関節筋が存在している.
  関節の自由度数$D$に対して, 必要最小限の筋は$D+1$本であるが, 大抵の筋骨格ヒューマノイドにはそれ以上の筋が配置されている.
  つまり,  軸駆動型ロボットで一つの関節が壊れるとそこで動作を停止するしかないのに対して, 一つの筋が切れたとしても, 他の筋によって動作を賄うことが可能である.
  筋は主に摩擦に強い合成繊維であるDyneemaによって構成されており, 可変剛性を可能とする非線形性弾性要素が筋と直列に配置されている場合が多い.
  ロボットによっては, モーメントアームを稼ぐためにpulleyで構成された筋経由点ユニットを使って筋を折り返している場合も存在する.
  柔軟な接触のために, 筋の周りには外装としての発泡材が巻かれている場合もあり, よりモデル化は困難を極める.
  それぞれの筋についてエンコーダから筋長$l$が・ロードセルから筋張力$f$・温度センサから筋温度$c$が測定できる.
  関節角度$\bm{\theta}$は球関節や複雑な肩甲骨ゆえに測定できない場合が多いが, 一部のロボットで測定することが可能である\cite{urata2006sensor, kawaharazuka2019musashi}.
  また, 視覚センサを用いることで, 実機関節角度を推定することができ, これを実機値として用いることも可能である\cite{kawaharazuka2018online}.
  これは以下のように, ロボットのエンドエフェクタにARマーカ等を取り付け, その位置$\bm{P}_{marker}$を取得し, 筋長変化からの関節角度推定値$\bm{\theta}^{est}$ \cite{ookubo2015learning}を初期値$\bm{\theta}^{init}$として$\bm{P}^{ref}=\bm{P}_{marker}$に対して逆運動学を解くという手法である.
  \begin{align}
    {\bm{\theta}^{est}}' = \textrm{IK}(\bm{P}^{ref}=\bm{P}_{marker}, \bm{\theta}^{init}=\bm{\theta}^{est})
  \end{align}
  ここで, IKは逆運動学を, ${\bm{\theta}^{est}}'$は視覚により補正された実機関節角度推定値となる.
  その他, 慣性センサや接触センサ等が存在する場合もある.
  注意として, 本研究は関節と筋の間の関係は学習するが, 関節の位置やリンク長に関する幾何モデルは正しいと仮定している.
}%

\begin{figure}[t]
  \centering
  \includegraphics[width=0.8\columnwidth]{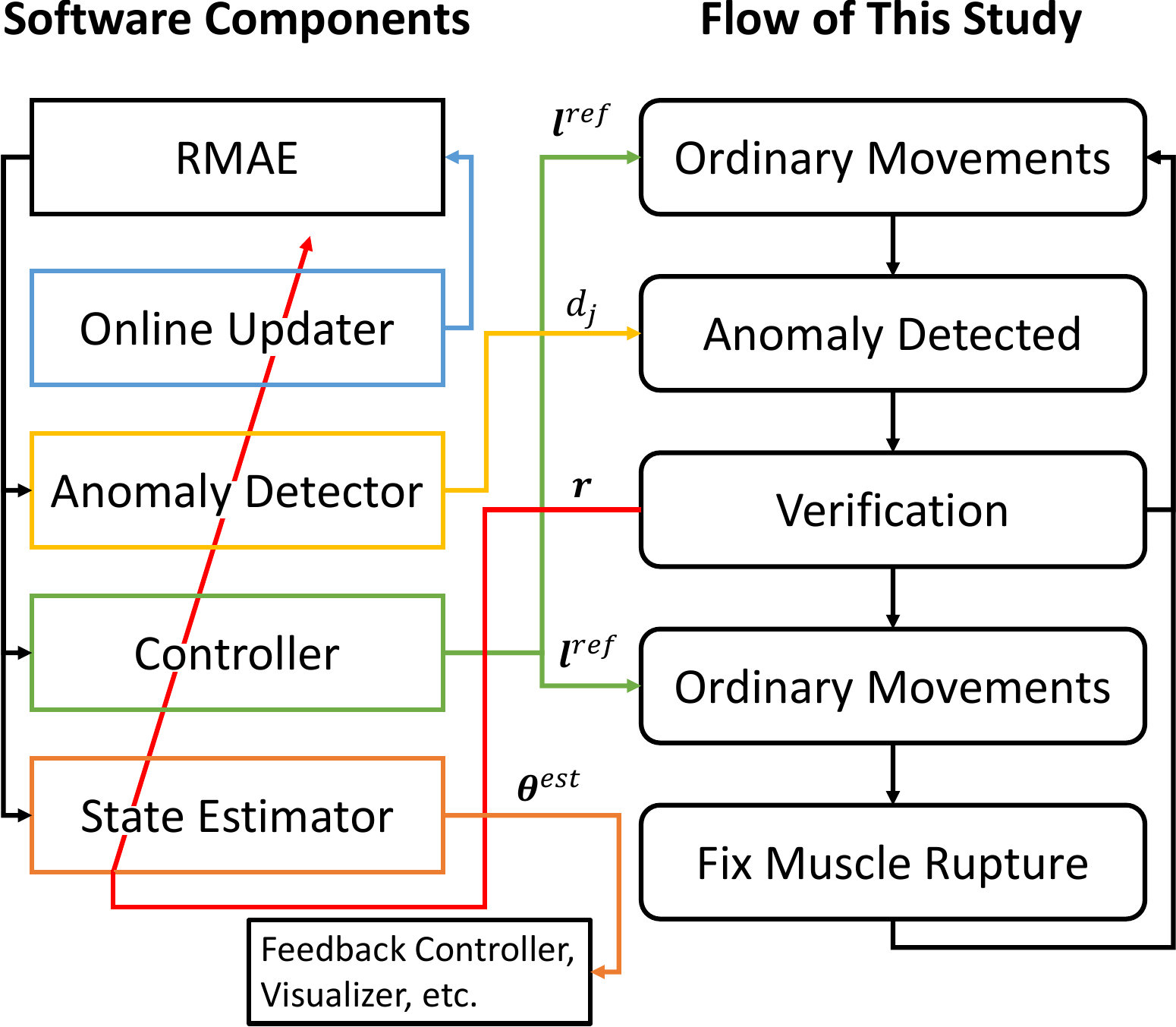}
  \caption{Overview of software components and flow of this study. The muscle rupture information $\bm{r}$ from Verification (red line) changes each component of online updater, anomaly detector, controller, and state estimator.}
  \label{figure:flow}
\end{figure}

\section{Utilization of Muscle Redundancy in Musculoskeletal Humanoids} \label{sec:proposed-methods}
\switchlanguage%
{%
  The overall flow of this study is shown in \figref{figure:flow}.
  The musculoskeletal humanoid with the flexible body is considered to have the following requirements for a robust motion strategy that allows it to continue to move even when one muscle is ruptured.
  \begin{itemize}
    \item The ability to detect muscle rupture.
    \item The ability to relearn the body model after the muscle rupture.
    \item Ease of muscle replacement.
  \end{itemize}
  First, it is necessary to be able to detect muscle rupture, but there can be various states of muscle rupture in musculoskeletal robots.
  In this study, muscle rupture is comprehensively defined as the state in which the muscle is broken due to friction or external force and the state in which the muscle cannot be moved due to malfunction of the circuit or motor, etc.
  Once the muscle rupture is detected, the motion is stopped, the state of the muscle is checked, and then the next motion strategy can be planned.
  Next, the relationship among muscle length, muscle tension, and joint angle in the flexible body should be relearned after the muscle rupture, because the relationship among them has been changed.
  Then, the body should be able to move as before the muscle rupture.
  Finally, if more than one muscle have been ruptured, the body will not be able to continue moving.
  Therefore, the muscles must be easily replaceable by humans.

  There are several other strategies that can be considered.
  The first strategy is to stop the robot until the muscles are replaced, without relearning the body model.
  This is a conventional approach, and while the algorithm is very simple, it does not use the robot to its full potential.
  The second strategy is to keep learning the body model without detecting the muscle rupture.
  This simplifies the algorithm as in the first strategy, but our experiments in \secref{sec:experiments} show that if the body model continues to be learned without including the muscle rupture information, the model changes significantly and the control and state estimation cannot be performed as intended.
  The third strategy is to self-heal muscles like humans without replacing muscles \cite{nakashima2019healing}.
  If this approach could be applied to this study, a complete system without human intervention could be constructed.
  However, the current technology has several problems, such as large unit size and large decrease in the transmissible tension after self-healing compared to before the muscle rupture.
  Therefore, in this study, we describe a strategy for robust continuous motion using muscle redundancy, based on the approach of muscle rupture detection, online learning of the body model including muscle rupture information, and subsequent muscle replacement.
}%
{%
  本研究の全体の流れを\figref{figure:flow}に示す.
  柔軟身体を持つ筋骨格ヒューマノイドが, 筋が切れても動き続けるロバストな動作戦略のために必要な要件は以下であると考える.
  \begin{itemize}
    \item 筋が切れたことが検知できる.
    \item 筋が切れた状態における身体センサ間関係を再学習する.
    \item 容易に筋を交換できる.
  \end{itemize}
  まず, 筋の破断を検知できる必要があるが, 一口に筋破断と言っても, ロボットにおける筋破断には様々な状態があり得る.
  本研究では, 筋破断や筋が切れたとは, ロボットにおいて, 摩擦や外部からの力によって筋が切れた状態, 回路やモータ等の故障により筋を能動的に動かせなくなった状態等, を総合的に表すものとする.
  筋の破断が検知できれば, 一旦動作を止め, 筋の状態を確認して, その後の動作戦略を練ることができる.
  次に, その筋が切れた状態で動作をする必要があるが, このとき, 柔軟な身体における筋長や筋張力, 関節の関係が変わってしまっているため, この関係を再学習する必要がある.
  これにより, 筋が切れる前の状態と同じように身体が動かせるようになるべきである.
  最後に, 筋が切れたのをそのままにしていては, さらに他の筋が切れた際に, 動作が継続できなくなってしまう.
  そのため, 最終的には人間が交換するが, この筋の交換が容易である必要がある.

  一方, この他にもいくつかの戦略が考えられる.
  一つ目は, 再学習等はせず, 筋を交換するまで動作を停止するという方法である.
  これは今までのアプローチであり, アルゴリズムが非常に単純になる一方, ロボットを動作限界まで使い切れていない.
  ニつ目は, 筋破断等は検知せず, 常に身体モデルを学習し続けるという方法である.
  この方法は一つ目同様アルゴリズムがシンプルになる一方, 筋破断に関する情報を含めずに身体モデルを学習し続けると, モデルが大きく変化し意図した通りに制御・状態推定ができなくなってしまうことが実験章から明らかになっている.
  三つ目は, 筋を交換せずとも人間のように自己修復するという方法である\cite{nakashima2019healing}.
  このアプローチを本研究に適用できれば人間が介入しない完全なシステムが構築可能である一方, 現状の技術ではユニットが大きい, 自己修復後に伝達可能な張力が破断前より大幅に減少する等の問題点がある.
  そのため, 本研究では筋破断検知, 筋破断情報を含めた身体センサ間関係のオンライン学習, その後の筋の交換というアプローチによる, 冗長性を使ったrobust continuous motion strategyに関する手法について述べる.
}%

\begin{figure}[t]
  \centering
  \includegraphics[width=0.8\columnwidth]{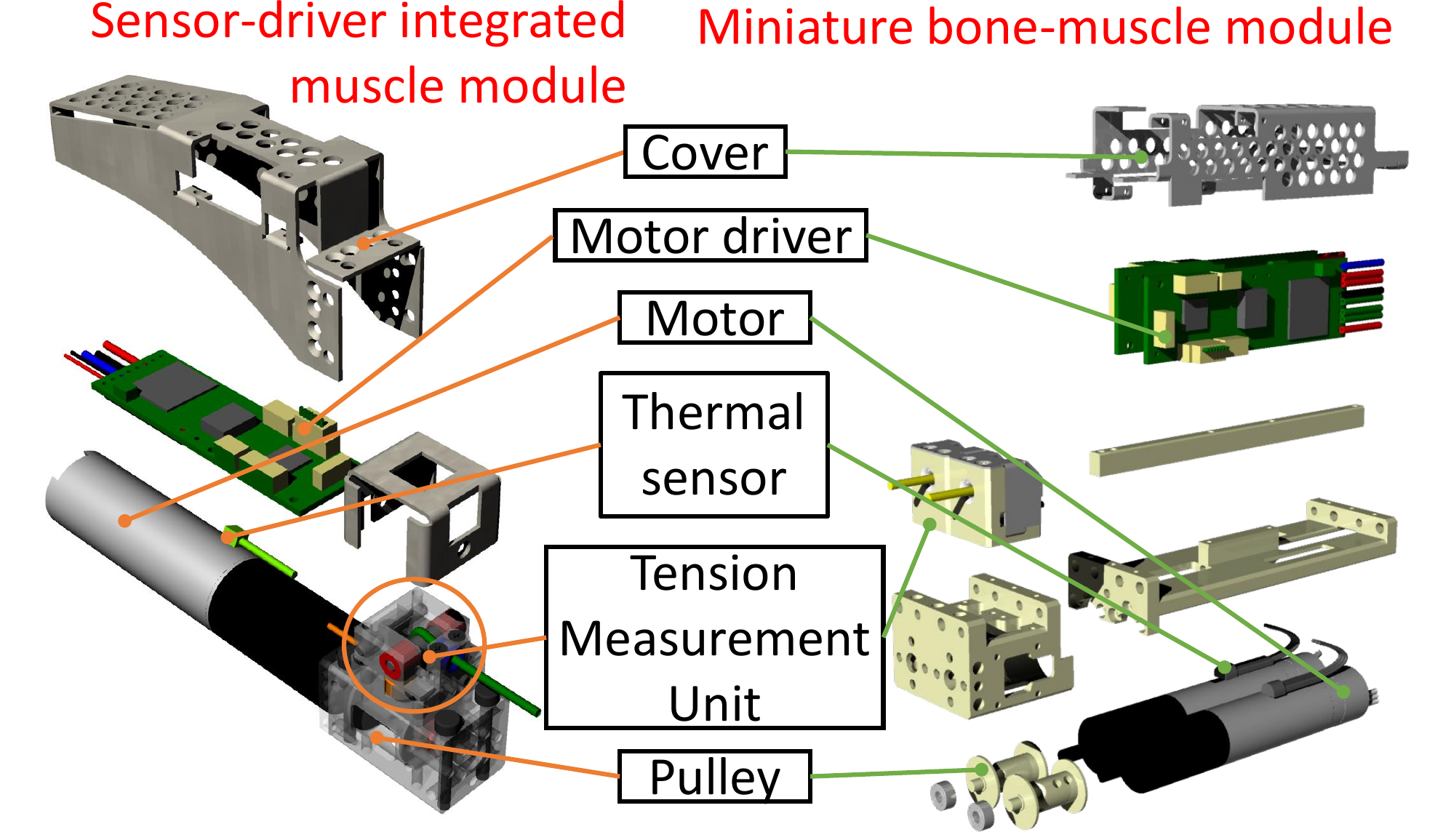}
  \caption{A sensor-driver integrated muscle module \cite{asano2015sensordriver} and a miniature bone-muscle module \cite{kawaharazuka2017forearm} for ease of replacement.}
  \label{figure:muscle-module}
\end{figure}

\subsection{Hardware and Software Requirements} \label{subsec:requirements}
\switchlanguage%
{%
  Regarding hardware, for the ease of muscle replacement, sensor-driver integrated muscle module \cite{asano2015sensordriver} and miniature bone-muscle module \cite{kawaharazuka2017forearm} have been developed (\figref{figure:muscle-module}).
  They are muscle modules which include motors, motor drivers, temperature sensors, and muscle tension measurement units in one package to improve reliability and maintainability.
  The previous muscle actuator \cite{kozuki2013muscle} has not been modularized and the motors and motor drivers have been located in different places, which requires a lot of effort to replace them.
  In contrast, for \cite{asano2015sensordriver, kawaharazuka2017forearm}, all that is required is to remove four screws, and replace and install the module again.
  From preliminary experiments, an experienced researcher was able to replace the whole module in about four minutes.
  In this study, the musculoskeletal humanoid Musashi \cite{kawaharazuka2019musashi}, which is equipped with these muscle modules, is used in the experiments.

  Regarding software, the main components required for this study are control, state estimation, and anomaly detection of the flexible body (state estimation is not directly necessary in this study, but is important in most cases for visualizer, feedback control, etc.).
  They are also updated online via a variable network structure such as a neural network and must be always adapted to the current body state.
  The muscle rupture cannot be completely determined only by detecting anomaly, so it is also important to verify the muscle rupture state afterwards.
  Finally, since some of the sensor values are improbable after the muscle rupture, all components such as online learning, anomaly detection, control, and state estimation should be modified accordingly.

}%
{%
  ハードウェアについては, 筋の交換容易性のため, sensor-driver integrated muscle module \cite{asano2015sensordriver}, miniature bone-muscle module \cite{kawaharazuka2017forearm}の2つの筋モジュールが開発されている(\figref{figure:muscle-module})
  これらは, モータ・モータドライバ・温度センサ・筋張力測定ユニット等の要素を一つのパッケージに収めることで信頼性・メンテナンス性を向上させた筋モジュールである.
  これまでは, 筋アクチュエータ\cite{kozuki2013muscle}はモジュール化されず, モータやモータドライバがそれぞれ別の場所にある形であり, 交換の際には多大な労力を要した.
  これに対して, \cite{asano2015sensordriver, kawaharazuka2017forearm}はネジを4つ外し, 交換し, また取り付けるのみであり, 非常に交換が簡単である.
  実験したところ, 熟練した研究者で, 約4分で交換可能であった.
  本研究では, これらの筋モジュールを搭載した筋骨格ヒューマノイドMusashi \cite{kawaharazuka2019musashi}を実験では用いている.

  ソフトウェアについては, 本研究を遂行するにあたり必要な要素は主に, 筋破断に関する異常検知と制御, 状態推定である(状態推定は本研究では直接は必要ないが, 大抵の場合他タスクを実行する上で重要である).
  また, これらはある可変なネットワーク構造(本研究ではRMAE)を介してオンラインで学習され, 常に現在の身体状態に適応していく必要がある.
  筋破断は異常検知をしただけでは完全には判断できないため, その後筋破断状態を確認する手順も重要である.
  最後に, 筋破断時には一部のセンサ値が通常あり得ない値を取ってしまうため, オンライン学習・異常検知・制御・状態推定の全コンポーネントをそれに応じて変化させる必要がある.

  以降では, \secref{subsec:rmae}で可変なネットワークであるRMAEの構造を, \secref{subsec:initial-training}でRMAEの初期学習を, \secref{subsec:online-learning}でRMAEのオンライン学習を, \secref{subsec:controller}でRMAEを用いた制御を, \secref{subsec:state-estimation}でRMAEを用いた状態推定を, \secref{subsec:detector}でRMAEを用いた異常検知を, \secref{subsec:verification}で筋破断状態の確認を述べていく.
}%

\begin{figure}[t]
  \centering
  \includegraphics[width=0.8\columnwidth]{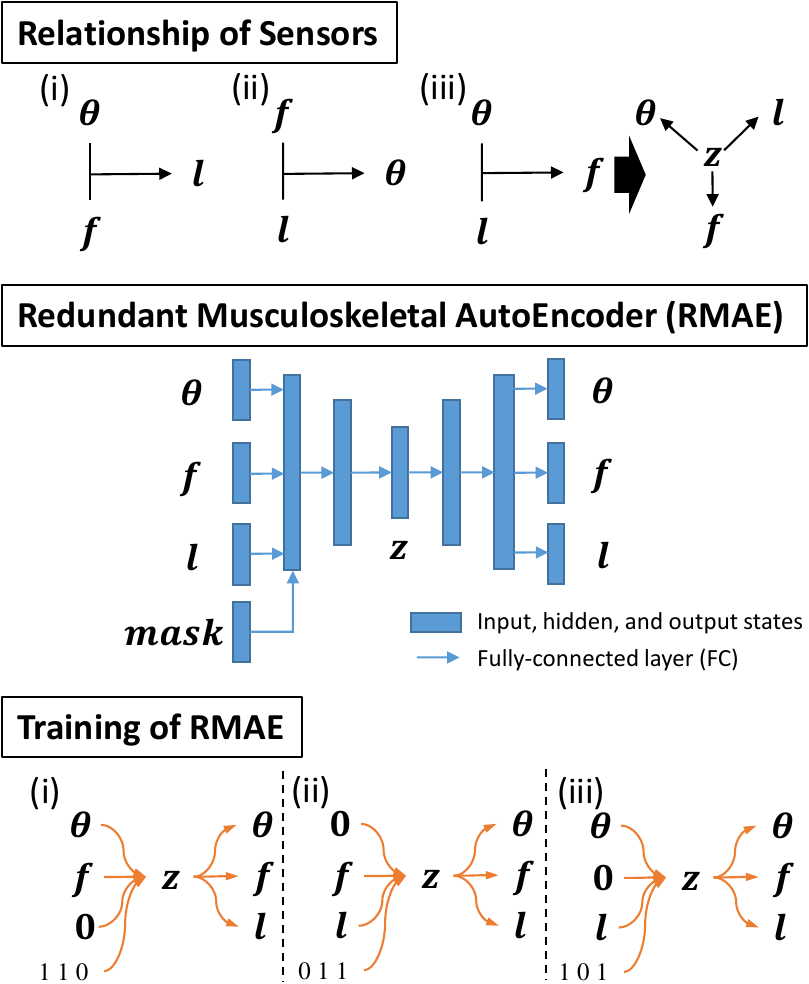}
  \caption{The relationship of sensors, the network structure of RMAE, and the training of RMAE.}
  \label{figure:network-structure}
\end{figure}

\begin{table}[t]
  \centering
  \caption{Notations in this article.}
  \vspace{0.1in}
  \scalebox{0.85}{
    \begin{tabular}{c|c}
      Notation & Definition \\\hline
      $D$ & the number of joints\\
      $M$ & the number of muscles\\
      $\bm{\theta}$ & joint angle $\in \mathcal{R}^{D}$\\
      $\bm{\theta}^{est}$ & the estimated joint angle $\in \mathcal{R}^{D}$\\
      $\bm{l}$ & muscle length $\in \mathcal{R}^{M}$\\
      $\bm{f}$ & muscle tension $\in \mathcal{R}^{M}$\\
      $\bm{c}$ & muscle temperature $\in \mathcal{R}^{M}$\\
      $\bm{mask}$ & the mask vector inputted in RMAE $\in \{0, 1\}^{3}$\\
      $\bm{z}$ & the latent value of RMAE $\in \mathcal{R}^{D+M}$\\
      $\bm{r}$ & the muscle rupture state $\in \{0, 1\}^{M}$\\
      $\bm{\tau}^{ref}$ & target joint torque $\in \mathcal{R}^{D}$\\
      $\bm{G}$ & muscle Jacobian $\in \mathcal{R}^{{M}\times{D}}$\\
      $\bm{h}_{\{enc, dec\}}$ & the encoder or decoder part of RMAE\\
      $\{\bm{\theta}, \bm{f}, \bm{l}\}^{data}$ & data of $\{\bm{\theta}, \bm{f}, \bm{l}\}$ accumulated for training of RMAE\\
      $\{\bm{\theta}, \bm{f}, \bm{l}\}^{cur}$ & the current measured sensor data of $\{\bm{\theta}, \bm{f}, \bm{l}\}$\\
      $\{\bm{\theta}, \bm{f}, \bm{l}\}^{pred}$ & $\{\bm{\theta}, \bm{f}, \bm{l}\}$ predicted from RMAE\\
      $\{\bm{\theta}, \bm{f}, \bm{l}\}^{ref}$ & target value of $\{\bm{\theta}, \bm{f}, \bm{l}\}$\\
      $\{\theta, f, l\}_{i}$ & the $i$-th value of $\{\bm{\theta}, \bm{f}, \bm{l}\}$
    \end{tabular}
  }
  \label{table:notations}
\end{table}

\subsection{Redundant Musculoskeletal Autoencoder} \label{subsec:rmae}
\begin{figure*}[t]
  \centering
  \includegraphics[width=1.99\columnwidth]{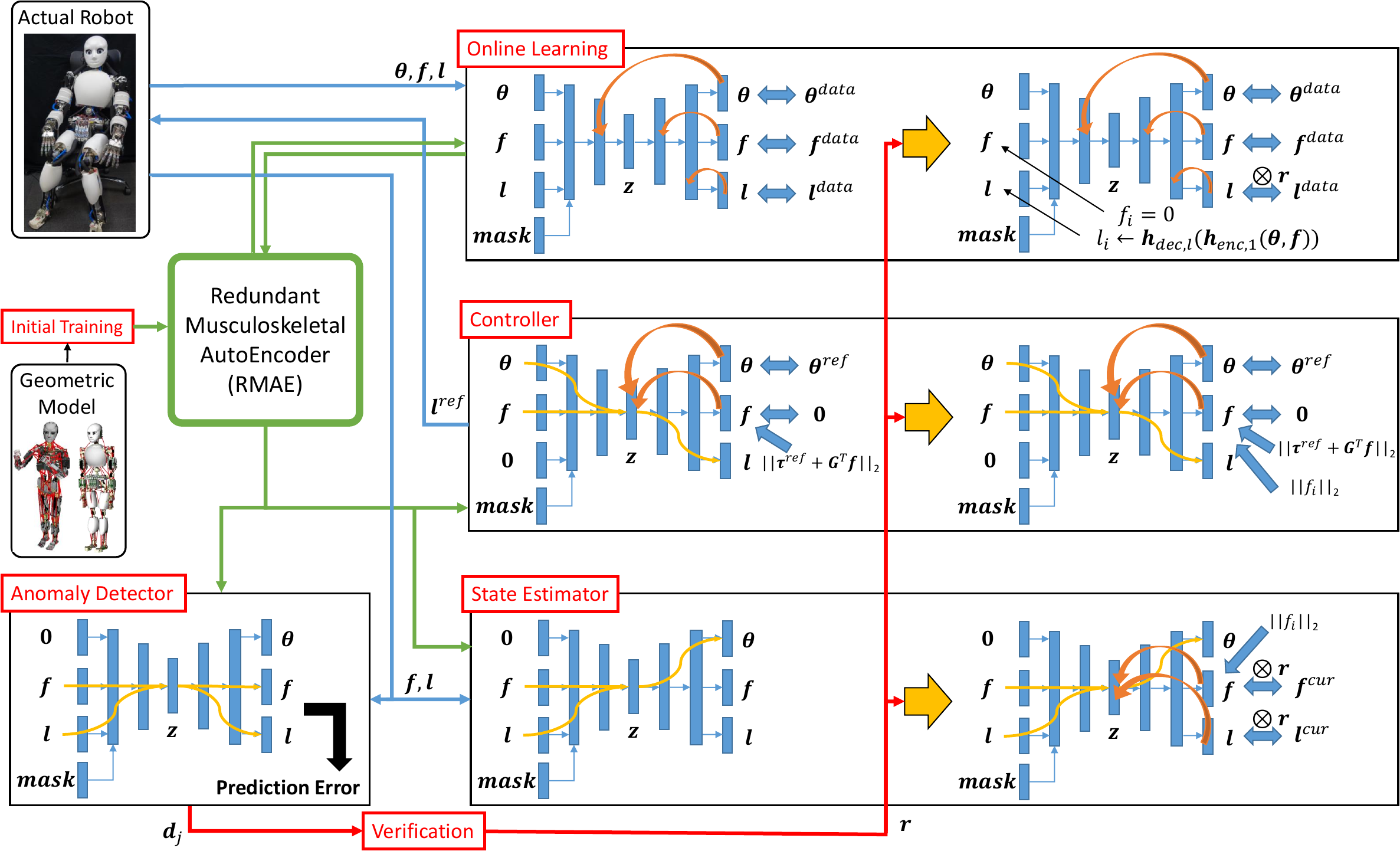}
  \caption{The detailed overall software system.}
  \label{figure:software-overview}
\end{figure*}

\switchlanguage%
{%
  The structure of RMAE (Redundant Musculoskeletal AutoEncoder), which is the core of this work, is shown in \figref{figure:network-structure}.
  The structure itself is equivalent to MAE of \cite{kawaharazuka2020autoencoder}, but the main technical novelty of this study is the addition of anomaly detection and muscle rupture verification using RMAE, and the changes based on muscle rupture information in online learning, control, and state estimation.
  The notations in this article are shown in \tabref{table:notations}.

  RMAE represents an intersensory relationship among redundant sensors, and we focus on the joint angle $\bm{\theta}$, muscle tension $\bm{f}$, and muscle length $\bm{l}$ that are usually obtained in musculoskeletal humanoids as shown in \secref{sec:musculoskeletal-structure}.
  There exist three relationships of (i) $(\bm{\theta}, \bm{f})\to\bm{l}$, (ii) $(\bm{f}, \bm{l})\to\bm{\theta}$, and (iii) $(\bm{l}, \bm{\theta})\to\bm{f}$, as shown in the upper diagram of \figref{figure:network-structure}.
  That is, among the three sensor values $(\bm{\theta}, \bm{f}, \bm{l})$, one can be inferred from the other two, which corresponds to the fact that all three sensor information can be obtained from two of the three sensors.
  The latent variable $\bm{z}$ can be calculated from two of the three sensors and $\bm{z}$ has information of all three sensors.

  Thus, condensing these three relationships into a single network, an AutoEncoder-type network \cite{hinton2006reducing} that inputs $(\bm{\theta}, \bm{f}, \bm{l})$ and $\bm{mask}$ and outputs the same $(\bm{\theta}, \bm{f}, \bm{l})$, as shown in the middle diagram of \figref{figure:network-structure}, is constructed.
  $\bm{mask}$ is a variable that switches between (i) -- (iii) in \figref{figure:network-structure}.
  Let $D$ be the number of joints and $M$ be the number of muscles, and the number of dimensions of $(\bm{\theta}, \bm{f}, \bm{l}, \bm{mask})$ are $(D, M, M, 3)$.
  Let $\bm{z}$ be the value of the middle layer with the smallest number of units in RMAE, with a dimension of $D+M$.
  This is because the minimum number of units can be smaller than $D+M$ since all values of $(\bm{\theta}, \bm{f}, \bm{l})$ can be inferred from $(\bm{\theta}, \bm{f})$.
  In this study, RMAE consists of seven fully-connected (FC) layers including the input and output layers, and the number of units is set to $(D+2M+3, 200, 30, D+M, 30, 200, D+2M)$.
  The numbers of units in the middle layers such as 30 and 200 are experimentally determined to be the minimum value that can explain the obtained data reasonably well.
  As the number of units increases, the accuracy improves, but the amount of computation increases, and over-fitting tends to occur.
  The activation function other than the last layer is hyperbolic tangent.
  The unit of $\bm{\theta}$ is [rad], the unit of $\bm{f}$ is [N/200], and the unit of $\bm{l}$ is [mm/100], in order to roughly adjust the scale.
  The current network structure of RMAE takes about 1.5 msec for a single inference and 3.0 msec for a single backpropagation, with the deep learning framework Chainer \cite{tokui2015chainer}.
  The algorithm itself consumes very little memory, while the network takes up about 10--100 Kbytes of memory depending on the number of input and output dimensions.

  Next, we describe the outline of the training method of RMAE.
  Although the structure of RMAE is similar to that of the usual AutoEncoder, the training method is special because of the presence of $\bm{mask}$.
  Using the fact that all three sensor information can be obtained from two of $(\bm{\theta}, \bm{f}, \bm{l})$, as shown in (i) in the lower diagram of \figref{figure:network-structure}, when the sensor information is narrowed down to $(\bm{\theta}, \bm{f})$, $(\bm{\theta}, \bm{f}, \bm{0}, \begin{pmatrix}1&1&0\end{pmatrix}^{T})$ is taken as input and $(\bm{\theta}, \bm{f}, \bm{l})$ is output.
  The value set to 0 in $\bm{mask}$ needs to be $\bm{0}$.
  (ii) and (iii) are run in the same way while changing the $\bm{mask}$, and the training is performed so that the input and output are close to each other.
  First, RMAE is initially trained from a geometric model including the relationship between joints and muscles.
  After that, the actual sensor data is obtained and this network is updated online.

  Some functions used in this study are defined as follows,
  \begin{align}
    \bm{z}&=\bm{h}_{enc}(\bm{\theta}, \bm{f}, \bm{l}, \bm{mask}) \label{eq:enc}\\
    \bm{z}&=\bm{h}_{enc, \textrm{i}}(\bm{\theta}, \bm{f})=\bm{h}_{enc}(\bm{\theta}, \bm{f}, \bm{0}, \begin{pmatrix}1&1&0\end{pmatrix}^{T}) \label{eq:enc-1}\\
    \bm{z}&=\bm{h}_{enc, \textrm{ii}}(\bm{f}, \bm{l})=\bm{h}_{enc}(\bm{0}, \bm{f}, \bm{l}, \begin{pmatrix}0&1&1\end{pmatrix}^{T}) \label{eq:enc-2}\\
    \bm{z}&=\bm{h}_{enc, \textrm{iii}}(\bm{\theta}, \bm{l})=\bm{h}_{enc}(\bm{\theta}, \bm{0}, \bm{l}, \begin{pmatrix}1&0&1\end{pmatrix}^{T}) \label{eq:enc-3}\\
    (\bm{\theta}, \bm{f}, \bm{l})&=\bm{h}_{dec}(\bm{z}) \label{eq:dec}\\
    \bm{\theta}&=\bm{h}_{dec, \bm{\theta}}(\bm{z}) \label{eq:dec-theta}\\
    \bm{f}&=\bm{h}_{dec, \bm{f}}(\bm{z}) \label{eq:dec-f}\\
    \bm{l}&=\bm{h}_{dec, \bm{l}}(\bm{z}) \label{eq:dec-l}
  \end{align}
  where $\bm{h}_{enc}$ is the encoder part of RMAE, and $\bm{h}_{enc, \{\textrm{i, ii, iii}\}}$ expresses each condition of encoder corresponding to (i), (ii), or (iii) in \figref{figure:network-structure}.
  Also, $\bm{h}_{dec}$ is the decoder part of RMAE, and $\bm{h}_{dec, \{\bm{\theta}, \bm{f}, \bm{l}\}}$ expresses each decoder outputting $\{\bm{\theta}, \bm{f}, \bm{l}\}$.

  A detailed overall software system surrounding RMAE is shown in \figref{figure:software-overview}.
  RMAE is at the center of this system and each of its components is executed by using backpropagation technique \cite{rumelhart1986backprop}.
}%
{%
  本研究の中核となる(Redundant Musculoskeletal AutoEncoder) RMAEの構造を\figref{figure:network-structure}に示す.
  この構造は\cite{kawaharazuka2020autoencoder}と同等のものであるが, 異常検知や筋破断確認が加わり, また, オンライン学習・制御・状態推定の筋破断情報に基づく変容が主な技術的新規性である.

  RMAEは冗長なセンサ群の相互関係を表現するが, \secref{sec:musculoskeletal-structure}に示した, 筋骨格ヒューマノイドで通常得られる関節角度$\bm{\theta}$, 筋張力$\bm{f}$, 筋長$\bm{l}$に着目する.
  これらの相互関係をニューラルネットワークにより可変かつ微分可能なな形で表現することで, その関係を更新したり制御や状態推定等に利用することができる.
  これらの間には, \figref{figure:network-structure}の上図に示すように, (i) $(\bm{\theta}, \bm{f})\to\bm{l}$, (ii) $(\bm{f}, \bm{l})\to\bm{\theta}$, (iii) $(\bm{l}, \bm{\theta})\to\bm{f}$という関係が存在する.
  つまり, $(\bm{\theta}, \bm{f}, \bm{l})$の3つのセンサ情報のうち, 2つから残り1つを求めることができる, 言い換えれば, $(\bm{\theta}, \bm{f}, \bm{l})$のうち2つのセンサ情報から, 3つ全てのセンサ情報を取得できることに相当する.
  3つのうち2つのセンサから潜在変数$\bm{z}$が計算でき, この$\bm{z}$が3つのセンサに関する情報を持っているのである.

  よってこの3つの関係を一つのネットワークに縮約すると, \figref{figure:network-structure}の中図に示すような, $(\bm{\theta}, \bm{f}, \bm{l})$と$\bm{mask}$を入力し, 同様の$(\bm{\theta}, \bm{f}, \bm{l})$を出力するAutoEncoder型\cite{hinton2006reducing}のネットワークとなる.
  このmaskは\figref{figure:network-structure}の(i)--(iii)を切り替える変数である.
  ここで, 関節数を$D$, 筋数を$M$とすると, $(\bm{\theta}, \bm{f}, \bm{l}, \bm{mask})$の次元数はそれぞれ$(D, M, M, 3)$である.
  RMAEにおいて最もユニット数の少ない中間層の値を$\bm{z}$とし, この次元は$D+M$とする.
  これは, $(\bm{\theta}, \bm{f})$から$(\bm{\theta}, \bm{f}, \bm{l})$の全ての値を推論できるため, 最小のユニット数は$D+M$よりも小さくすることが可能であるからである.
  本研究ではRMAEは入力と出力層を含めた7層のfully connected layer (FC)とし, そのユニット数は$(D+2M+3, 200, 30, D+M, 30, 200, D+2M)$とし, 活性化関数はhyperbolic tangentとしている.
  また, スケールを大まかに揃えるため, $\bm{\theta}$の単位は[rad], $\bm{f}$の単位は[N/200], $\bm{l}$の単位は[mm/100]としている.

  次に, RMAE学習方法の概要について述べる.
  RMAEは通常のAutoEncoderと構造は似ているものの, $\bm{mask}$があるため学習方法は特殊である.
  $(\bm{\theta}, \bm{f}, \bm{l})$のうち2つのセンサ情報から3つのセンサ情報全てを取得できるという性質を用いて, \figref{figure:network-structure}下図の(i)のように, 得られるセンサ情報を$(\bm{\theta}, \bm{f})$に絞った場合は$(\bm{\theta}, \bm{f}, \bm{0}, \begin{pmatrix}1&1&0\end{pmatrix}^{T})$を入力として, $(\bm{\theta}, \bm{f}, \bm{l})$を出力する.
  つまり, $mask$で0とした値を$\bm{0}$にする必要がある.
  (ii), (iii)も$\bm{mask}$を変更させながら同様に実行し, 入力と出力が近くなるように学習を行う.
  まず関節と筋の関係を含む幾何モデルからRMAEを初期学習させる.
  その後, 実機センサデータを取得し, このネットワークをオンラインで更新していくことになる.

  本研究で使用するいくつかの関数を以下のように定義する.
  \begin{align}
    \bm{z}=\bm{h}_{enc}(\bm{\theta}, \bm{f}, \bm{l}, \bm{mask}) \label{eq:enc}\\
    \bm{z}&=\bm{h}_{enc, \textrm{i}}(\bm{\theta}, \bm{f})=\bm{h}_{enc}(\bm{\theta}, \bm{f}, \bm{0}, \begin{pmatrix}1&1&0\end{pmatrix}^{T}) \label{eq:enc-1}\\
    \bm{z}&=\bm{h}_{enc, \textrm{ii}}(\bm{f}, \bm{l})=\bm{h}_{enc}(\bm{0}, \bm{f}, \bm{l}, \begin{pmatrix}0&1&1\end{pmatrix}^{T}) \label{eq:enc-2}\\
    \bm{z}&=\bm{h}_{enc, \textrm{iii}}(\bm{\theta}, \bm{l})=\bm{h}_{enc}(\bm{\theta}, \bm{0}, \bm{l}, \begin{pmatrix}1&0&1\end{pmatrix}^{T}) \label{eq:enc-3}\\
    (\bm{\theta}, \bm{f}, \bm{l}) = \bm{h}_{dec}(\bm{z}) \label{eq:dec}\\
    \bm{\theta} = \bm{h}_{dec, \bm{\theta}}(\bm{z}) \label{eq:dec-theta}\\
    \bm{f} = \bm{h}_{dec, \bm{f}}(\bm{z}) \label{eq:dec-f}\\
    \bm{l} = \bm{h}_{dec, \bm{l}}(\bm{z}) \label{eq:dec-l}
  \end{align}
  ここで, $\bm{h}_{enc}$をRMAEのencoderを表し, $\bm{h}_{enc, \{\textrm{i, ii, iii}\}}$はそれぞれ(i), (ii), (iii)に対応したencoderを表す.
  同様に, $\bm{h}_{dec}$をRMAEのdecoderを表し, $\bm{h}_{dec, \{\bm{\theta}, \bm{f}, \bm{l}\}}$はそれぞれ$\{\bm{\theta}, \bm{f}, \bm{l}\}$を出力するdecoderを表す.

  RMAEを取り巻くソフトウェアの詳細な全体システムを\figref{figure:software-overview}に示す.
  このRMAEを中心として誤差逆伝播を利用し, それぞれのコンポーネントが動作している.
}%

\subsection{Initial Training of RMAE} \label{subsec:initial-training}
\switchlanguage%
{%
  In order to update RMAE online, it is necessary to initialize RMAE with a geometric model of the musculoskeletal structure.
  The geometric model represents each muscle route by linearly connecting the start, relay, and end points of the muscle.
  The relationship between muscle length change and muscle tension in a nonlinear elastic element is identified and combined with the geometric model to obtain the data on $(\bm{\theta}, \bm{f}, \bm{l})$ while randomizing $\bm{\theta}$ and $\bm{f}$ in the geometric model.

  Using these data, (i), (ii), and (iii) in \figref{figure:network-structure} are randomly executed.
  In other words, one of $(\bm{\theta}, \bm{f}, \bm{l})$ is randomly set to $\bm{0}$, the data including the corresponding $\bm{mask}$ is inputted to RMAE, and the predicted data $(\bm{\theta}^{pred}, \bm{f}^{pred}, \bm{l}^{pred})$ is outputted.
  Finally, RMAE is trained by calculating the loss function $L_{initial}$ for $\{\bm{\theta}, \bm{f}, \bm{l}\}$ as follows,
  \begin{align}
    L_{initial} = w_{\theta}||\bm{\theta}-\bm{\theta}^{pred}||_{2} + w_{f}||\bm{f}-\bm{f}^{pred}||_{2}\\\nonumber+ w_{l}||\bm{l}-\bm{l}^{pred}||_{2}
  \end{align}
  where $||\cdot||_{2}$ represents L2 norm.

  In this study, the number of batches for initial training is set to $N^{init}_{batch}=100$, the number of epochs is set to $N^{init}_{epoch}=100$, $w_{\theta}=1.0$, $w_{f}=10.0$, $w_{l}=100.0$, and the optimization method is Adam \cite{kingma2015adam}.
  Various values were tried for these parameters, and they are adjusted to make each error for $\{\theta, f, l\}$ less than about 0.01.
  The model of the epoch with the smallest test loss was used.
  For the activation function and the optimization method, we tried various types and selected the algorithm that converges to the smallest loss.
}%
{%
  RMAEをオンラインで更新するためにはまず, 筋骨格構造の幾何モデルを用いてRMAEを初期学習させる必要がある.
  ここで言う幾何モデルとは, 筋の起始点・中継点・終止点を直線で結び筋経路を表現したモデルである.
  このモデルを用いると, 経由点間の距離から関節角度を指定した際の筋長を求めることができる.
  ここで, 幾何モデルから筋の絶対長さを求める関数を$\bm{h}_{geo, abs}(\bm{\theta})$, 初期姿勢(全関節角度が0, つまり$\bm{\theta}=\bm{0}$)からの相対筋長を求める関数を$\bm{h}_{geo}(\bm{\theta})$とする.

  まず, 筋単体のテストベッドにより, 筋長の全体長さ$\bm{l}_{abs}$に比例するダイニーマの伸びを考慮した, 筋張力$\bm{f}$と非線形弾性要素を含む筋長変化の関係$\Delta\bm{l}_{e}(\bm{l}_{abs}, \bm{f})$を関数フィッティングにより求める.
  次に, そのモデルを使い, 以下のようにランダムな関節角度$\bm{\theta}^{r}$とランダムな筋張力$\bm{f}^{r}$から擬似的にデータを生成する.
  \begin{align}
    (\bm{\theta}, \bm{f}, \bm{l}) = (\bm{\theta}^{r}, \bm{f}^{r}, \bm{h}_{geo}(\bm{\theta}^{r})-\Delta\bm{l}_{e}(\bm{h}_{geo, abs}(\bm{\theta}^{r}), \bm{f}^{r}))
  \end{align}
  このデータを用いて, \figref{figure:network-structure}における(i) (ii) (iii)をランダムに実行する.
  つまり, $(\bm{\theta}, \bm{f}, \bm{l})$のうち一つをランダムに$\bm{0}$にし, 対応する$\bm{mask}$を加えたデータをRMAEに入力し, $(\bm{\theta}^{pred}, \bm{f}^{pred}, \bm{l}^{pred})$を出力する.
  最後に, 損失関数$L_{initial}$を$\{\bm{\theta}, \bm{f}, \bm{l}\}$について以下のように計算し, RMAEを学習させる.
  \begin{align}
    L_{initial} = w_{\theta}||\bm{\theta}-\bm{\theta}^{pred}||_{2} + w_{f}||\bm{f}-\bm{f}^{pred}||_{2}\\\nonumber+ w_{l}||\bm{l}-\bm{l}^{pred}||_{2}
  \end{align}
  ここで, $||\cdot||_{2}$はL2ノルムを表す.

  本研究では学習の際のバッチ数を$N^{init}_{batch}=100$, エポック数を$N^{init}_{epoch}=30$, $w_{\theta}=1.0$, $w_{f}=10.0$, $w_{l}=100.0$とし, 最適化手法はAdam \cite{kingma2015adam}を用いる.
  それぞれのパラメータについては様々な値を試したが, $\{\theta, f, l\}$に関する誤差がそれぞれ0.01以下程度になるように調整しており, 最もtest lossが小さかったepochのmodelを使用している.
}%

\subsection{Online Learning of RMAE} \label{subsec:online-learning}
\switchlanguage%
{%
  We describe the online learning of RMAE using the sensor information of the actual robot.

  First, the current sensor data $(\bm{\theta}^{cur}, \bm{f}^{cur}, \bm{l}^{cur})$ is obtained from the actual robot.
  Note that the training data is generated when the robot is stationary and the joint angle or the muscle tension is farther than a certain threshold from the previous training data.

  Next, this data is accumulated and augmented, and RMAE is updated online.
  The data $(\bm{\theta}^{cur}, \bm{f}^{cur}, \bm{l}^{cur})$ is accumulated and when the number of data exceeds $N^{online}_{thre}$, the online update is started.
 $N^{online}_{data}$ ($N^{online}_{data}\leq{N}^{online}_{thre}$) random data from the accumulated data and the latest data are obtained.
  Also, one data point of $(\bm{0}, \bm{0}, \bm{0})$ is added.
  To these data, three $\bm{mask}$ of (i), (ii), and (iii) are added, respectively, and a total of $3\times(N^{online}_{data}+2)$ data is used to update the network in the same way with \secref{subsec:initial-training}, with the loss function $L_{online}$ as $L_{initial}$.
  Since the initial training builds up the overall network structure, the network smoothly matches the obtained data by online learning without destroying the structure.

  Here, it is necessary to change the online learning method depending on the muscle rupture information $\bm{r}$ (an $M$-dimensional vector with 0 for ruptured muscles and 1 for unruptured muscles).
  The changes in online learning, state estimation, and control based on $\bm{r}$ are the most important technical novelties of this study that take advantage of the muscle redundancy.
  When a muscle is ruptured, $f$ is always close to zero and $l$ does not change at all, so if the online learning continues as it is, the structure of RMAE will be severely destroyed.
  So first, for the obtained dataset $(\bm{\theta}, \bm{f}, \bm{l})$, let $\bm{f}$ of the ruptured muscle $i$, $f_i$, be 0.
  Next, since $\bm{l}$ of the ruptured muscle $i$, $l_i$, cannot be directly used, $\bm{l}^{pred}$ is inferred as $\bm{l}^{pred}=\bm{h}_{dec, \bm{l}}(\bm{h}_{enc, \textrm{i}}(\bm{\theta}, \bm{f}))$.
  Finally, $l_i$ is replaced with $l^{pred}_{i}$ ($\bm{l}^{pred}$ of the muscle $i$), and online learning is executed.
  Here, the loss related to $\bm{l}$ of muscle $i$ is not meaningful and must be removed from the loss function.
  Therefore, the loss function is written as follows,
  \begin{align}
    L'_{online} = &w_{\theta}||\bm{\theta}-\bm{\theta}^{pred}||_{2}\nonumber\\+&w_{f}||\bm{f}-\bm{f}^{pred}||_{2} + w_{l}||\bm{r}\otimes(\bm{l}-\bm{l}^{pred})||_{2}
  \end{align}
  where $\otimes$ expresses an element-wise product.

  In this study, the number of batches for online learning is set to $N^{online}_{batch}=10$, the number of epochs is set to $N^{online}_{epoch}=10$, and other constants are set to $N^{online}_{thre}=10$ and $N^{online}_{data}=10$.
  $N^{online}_{epoch}$ should be set much smaller than $N^{init}_{epoch}$, because if $N^{online}_{epoch}$ is too large, RMAE will fit the obtained data too well and over-fitting will occur.
}%
{%
  実機センサデータを用いたオンライン学習について述べる.

  まず, 実機からセンサデータ$(\bm{\theta}^{cur}, \bm{f}^{cur}, \bm{l}^{cur})$を取得する.
  ただし, これは関節角度か筋張力が前回の学習の際のデータからある閾値以上離れ, かつ静止している場合に学習データを作成する.

  次に, このデータを蓄積・拡張し, RMAEをオンラインで更新していく.
  データ$(\bm{\theta}^{cur}, \bm{f}^{cur}, \bm{l}^{cur})$を蓄積していき, データ数が$N^{online}_{thre}$を超えたところで, 更新を開始する.
  蓄積したデータの中から$N^{online}_{data}$ ($N^{online}_{data}\leq{N}^{online}_{thre}$)個, そして最新のデータ1個を取得する.
  また, $(\bm{0}, \bm{0}, \bm{0})$というデータも一つ加える.
  これらデータに対して, それぞれ(i) (ii) (iii)の3種類の$\bm{mask}$を加え, 計$3\times(N^{online}_{data}+2)$個のデータを用いて, 損失関数$L_{online}=L_{initial}$として\secref{subsec:initial-training}と同様にネットワークを更新する.
  初期学習により全体的なネットワーク構造が構築されるため, オンライン学習ではその構造を崩さずに得られたデータに滑らかに合致するようになる.

  ここで, 筋破断に関する情報$\bm{r}$ (切れた筋を0, 切れていない筋を1とした$M$次元ベクトル)に対して, オンライン学習の方法を変化させる必要がある.
  筋が切れた場合はその筋の$f$は常に0付近となり, $l$もほとんど変化しなくなるため, そのまま学習を続けると, 関係性が大きく崩れてRMAEの構造が破壊されてしまう.
  そこでまず, 得られたデータセット$(\bm{\theta}, \bm{f}, \bm{l})$に対して, 切れた筋$i$の$\bm{f}$, $f_i$を全て0とする.
  次に, 切れた筋$i$の$\bm{l}$である$l_i$を直接使うわけにはいかないため, $\bm{l}^{pred}=\bm{h}_{dec, \bm{l}}(\bm{h}_{enc, \textrm{i}}(\bm{\theta}, \bm{f}))$のように$\bm{l}^{pred}$を推論する.
  最後に, $l_i$を$l^{pred}_{i}$に置き換え, オンライン学習を実行させる.
  このとき, 筋$i$の$\bm{l}$に関する損失は意味を成さないため, 損失関数から抜く必要がある.
  よって, 損失関数は以下のようになる.
  \begin{align}
    L'_{online} = &w_{\theta}||\bm{\theta}-\bm{\theta}^{pred}||_{2}\nonumber\\+&w_{f}||\bm{f}-\bm{f}^{pred}||_{2} + w_{l}||\bm{r}\otimes(\bm{l}-\bm{l}^{pred})||_{2}
  \end{align}
  ここで, $\otimes$は要素ごとの掛け算を表す.

  本研究では, 更新の際のバッチ数を$N^{online}_{batch}=10$, エポック数を$N^{online}_{epoch}=10$とし, その他定数は$N^{online}_{thre}=10$, $N^{online}_{data}=10$とした.
  $N^{online}_{epoch}$は大きすぎると得られたデータに適合し過ぎてしまうため, initial trainingに比べずっと小さく設定する必要がある.
}%

\subsection{Controller Using RMAE} \label{subsec:controller}
\switchlanguage%
{%
  First, in this study, muscles of the musculoskeletal humanoid are controlled by the following muscle stiffness control \cite{shirai2011stiffness},
  \begin{align}
    \bm{f}^{send} = \bm{f}^{bias} + \textrm{max}(\bm{0}, K_{stiff}(\bm{l}^{cur}-\bm{l}^{ref})) \label{eq:muscle-stiffness-control}
  \end{align}
  where $\bm{f}^{send}$ is a muscle tension to be exerted, $\bm{f}^{bias}$ is a bias of muscle tension, and $K_{stiff}$ is a muscle stiffness coefficient.

  We describe a control method using RMAE.
  First, from the current muscle tension $\bm{f}^{cur}$ and target joint angle $\bm{\theta}^{ref}$, the latent state is calculated as $\bm{z}=\bm{h}_{enc, \textrm{i}}(\bm{\theta}^{ref}, \bm{f}^{cur})$.
  Then, the following process is executed repeatedly.
  \begin{enumerate}
    \renewcommand{\labelenumi}{\arabic{enumi})}
    \item Calculate $(\bm{\theta}^{pred}, \bm{f}^{pred}, \bm{l}^{pred}) = \bm{h}_{dec}(\bm{z})$.
    \item Calculate the loss function $L_{control}(\bm{\theta}^{pred}, \bm{f}^{pred}, \bm{l}^{pred})$.
    \item Update $\bm{z}$ using a backpropagation technique \cite{rumelhart1986backprop}.
  \end{enumerate}
  In 2), the loss function $L_{control}$ considering minimization of muscle tension and realization of target joint angle and necessary joint torque is calculated as below,
  \begin{align}
    L_{control}(\bm{\theta}, \bm{f}, \bm{l}) = &w_{1}||\bm{f}||_{2} + w_{2}||\bm{\theta}-\bm{\theta}^{ref}||_{2}\nonumber\\ &+ w_{3}||\bm{\tau}^{ref}+G^{T}(\bm{\theta}^{ref}, \bm{f}^{cur})\bm{f}||_{2} \label{eq:keep-control-loss}
  \end{align}
  where $w_{\{1, 2, 3\}}$ is a constant weight, and $\bm{\tau}^{ref}$ is a joint torque required to keep the posture of $\bm{\theta}^{ref}$, which is calculated from a geometric model.
  $G(\bm{\theta}, \bm{f})$ is a muscle Jacobian, which is calculated from the difference of the outputs of $\bm{l}$ when inputting $(\bm{\theta}$, $\bm{f})$ and $(\bm{\theta}+\Delta\bm{\theta}, \bm{f})$ to RMAE in the form of (i) ($\Delta\bm{\theta}$ is a small displacement of joint angle).
  In 3), $\bm{z}$ is updated by a gradient descent with a backpropagation technique \cite{rumelhart1986backprop} as below,
  \begin{align}
    \bm{g} &= \partial{L}_{control}/\partial\bm{z}\\
    \bm{z} &= \bm{z} - \gamma\bm{g}/|\bm{g}|
  \end{align}
  where $\bm{g}$ is a gradient of $L_{control}$ for $\bm{z}$, and $\gamma$ is a learning rate.
  Although $\gamma$ can be constant, batches are constructed with various $\gamma$ in this study, and the best $\gamma$ is selected at each step to achieve faster convergence.
  The maximum value of $\gamma$, $\gamma^{control}_{max}$, is determined, $[0, \gamma^{control}_{max}]$ is divided equally into $N^{control}_{batch}$ parts, and $N^{control}_{batch}$ batches of $\bm{z}$ updated by each learning rate $\gamma$ are created.
  1) and 2) are executed again, and $\bm{z}$ with the lowest $L_{control}$ is chosen.
  Steps 1)-3) are performed $N^{control}_{epoch}$ times.
  Finally, $\bm{l}^{ref}$ considering muscle stiffness control of \equref{eq:muscle-stiffness-control} is calculated as below using $\bm{l}^{pred}$ and $\bm{f}^{pred}$ obtained from $\bm{z}$, and is sent to the robot.
  \begin{align}
    \bm{l}^{comp}(\bm{f}) = -(\bm{f} - \bm{f}^{bias})/K_{stiff} \\
    \bm{l}^{ref} = \bm{l}^{pred}+\bm{l}^{comp}(\bm{f}^{pred}) \label{eq:move-keep}
  \end{align}

  This makes it possible to find $\bm{l}^{ref}$ such that muscle tension is minimized in realizing the joint angle $\bm{\theta}^{ref}$.
  Although it is possible to directly calculate necessary muscle tension $\bm{f}^{ref}$ by solving a quadratic programming minimizing $||\bm{f}||_{2}$ and satisfying $\bm{\tau}^{ref}=-G^T\bm{f}$, there is no guarantee that $\bm{f}^{ref}$ can be realized at $\bm{\theta}^{ref}$.
  Therefore, in this study, $\bm{f}^{ref}$ is searched for in the latent space $\bm{z}$.

  Also, it is necessary to modify the control method for the muscle rupture information $\bm{r}$.
  Here, the muscle tension of the ruptured muscle $i$, $f_i$, must be zero.
  Therefore, \equref{eq:keep-control-loss} is partially modified to calculate the following loss,
  \begin{align}
    L'_{control}(\bm{\theta}, \bm{f}, \bm{l}) = L_{control}(\bm{\theta}, \bm{f}, \bm{l}) + w_{4}||f_i||_{2}
  \end{align}
  where $w_{4}$ is a constant weight of which the value should be much larger than $w_{1}$.

  In this study, we set $w_{1}=1.0$, $w_{2}=1.0$, $w_{3}=0.01$, $w_{4}=10.0$, $\gamma^{control}_{max}=0.5$, $N^{control}_{batch}=10$, and $N^{control}_{epoch}=10$.
}%
{%
  前提として, 本研究では筋骨格ヒューマノイドを以下のような筋剛性制御\cite{shirai2011stiffness}により動作させる.
  \begin{align}
    \bm{f}^{ref} = \bm{f}^{bias} + \textrm{max}(\bm{0}, K_{stiff}(\bm{l}^{cur}-\bm{l}^{ref})) \label{eq:muscle-stiffness-control}
  \end{align}
  ここで, $\bm{f}^{ref}$は指令筋張力, $\bm{f}^{bias}$は筋剛性制御のバイアス項, $K_{stiff}$は筋剛性制御の筋剛性係数である.
  筋剛性の値は本研究ではある一定値としているが, 大きくし過ぎると筋長が現在値と指令値の誤差を許容できなくなってしまい, 低すぎると追従性が悪くなる.

  RMAEを用いた制御手法について説明する.
  まず, 現在の筋張力$\bm{f}^{cur}$, 指令関節角度$\bm{\theta}^{ref}$から, 潜在状態$\bm{z}=\bm{h}_{enc, \textrm{i}}(\bm{\theta}^{ref}, \bm{f}^{cur})$を求める.
  次に以下の工程を繰り返す.
  \begin{enumerate}
    \renewcommand{\labelenumi}{\arabic{enumi})}
    \item $(\bm{\theta}^{pred}, \bm{f}^{pred}, \bm{l}^{pred}) = \bm{h}_{dec}(\bm{z})$を求める.
    \item 損失$L_{control}(\bm{\theta}^{pred}, \bm{f}^{pred}, \bm{l}^{pred})$を計算する.
    \item 誤差逆伝播\cite{rumelhart1986backprop}により$\bm{z}$を更新する.
  \end{enumerate}
  2)では, 以下のように, 筋張力最小化・指令関節角度実現・必要関節トルク実現に関する損失$L_{control}$を計算する.
  \begin{align}
    L_{control}(\bm{\theta}, \bm{f}, \bm{l}) = &w_{1}||\bm{f}||_{2} + w_{2}||\bm{\theta}-\bm{\theta}^{ref}||_{2}\nonumber\\ &+ w_{3}||\bm{\tau}^{ref}+G^{T}(\bm{\theta}^{ref}, \bm{f}^{cur})\bm{f}||_{2} \label{eq:keep-control-loss}
  \end{align}
  ここで, $\bm{\tau}^{ref}$は幾何モデルから計算された$\bm{\theta}^{ref}$を保つのに必要な関節トルク値である.
  また, $G(\bm{\theta}, \bm{f})$は筋長ヤコビアンであり, 微小な変位$\Delta\bm{\theta}$を加えた$(\bm{\theta}+\Delta\bm{\theta}$, $\bm{f})$, または$(\bm{\theta}, \bm{f})$を(i)の形でRMAEに入力したときの出力$\bm{l}$の差から計算することができる.
  3)では, 以下のように誤差逆伝播法\cite{rumelhart1986backprop}を元に$\bm{z}$を最急降下法で更新する.
  \begin{align}
    \bm{g} &= \partial{L}_{control}/\partial\bm{z}\\
    \bm{z} &= \bm{z} - \gamma\bm{g}/|\bm{g}|
  \end{align}
  ここで, $\bm{g}$は$\bm{z}$に関する$L_{control}$の勾配, $\gamma$は学習率を表す.
  このとき, $\gamma$の値を決め打ちしても良いが, 本研究では様々な$\gamma$によってバッチを作成し, 最も良い$\gamma$を選ぶことでより速い収束を目指す.
  $\gamma$の最大値$\gamma^{control}_{max}$を決め, 0から$\gamma^{control}_{max}$までの値を$N^{control}_{batch}$等分し, それぞれの学習率によって更新された$\bm{z}$を$N^{control}_{batch}$個作成する.
  もう一度1)と2)を行い, 最も$L_{control}$が小さかった$\bm{z}$を採用する.
  これら1)--3)の工程を, $N^{control}_{epoch}$回行う.
  そして最後に, $\bm{z}$から得られた$\bm{l}^{pred}$, $\bm{f}^{pred}$を用いて筋剛性制御を考慮したうえで以下のように$\bm{l}^{ref}$を計算し, 実機に送る.
  \begin{align}
    \bm{l}^{comp}(\bm{f}) = -(\bm{f} - \bm{f}^{bias})/K_{stiff} \\
    \bm{l}^{ref} = \bm{l}^{pred}+\bm{l}^{comp}(\bm{f}^{pred}) \label{eq:move-keep}
  \end{align}
  これにより, 関節角度$\bm{\theta}^{ref}$を実現する中で, 筋張力が最小となるような$\bm{l}^{ref}$を求めることが可能となる.
  $||\bm{f}||_{2}$を最小化し$\bm{\tau}^{ref}=-G^T\bm{f}$を満たすような二次計画法を解いて直接必要な筋張力$\bm{f}^{ref}$を求めることも可能だが, $\bm{\theta}^{ref}$において$\bm{f}^{ref}$を実現できる保証はなく, 本研究では潜在空間$\bm{z}$内において$\bm{f}^{ref}$を探索している.

  また, 筋が切れたという情報$\bm{r}$に対して, 制御手法を変化させる必要がある.
  この場合, 切れた筋$i$の筋張力$f_i$は必ず0にならなければならない.
  ゆえに, \equref{eq:keep-control-loss}を一部変更し, 以下のようなlossを計算する.
  \begin{align}
    L'_{control}(\bm{\theta}, \bm{f}, \bm{l}) = L_{control}(\bm{\theta}, \bm{f}, \bm{l}) + w_{4}||f_i||_{2}
  \end{align}
  ここで, $w_{4}$は係数であり, この値は$w_{1}$に比べて十分大きくする必要がある.

  本研究では, $w_{1}=1.0$, $w_{2}=1.0$, $w_{3}=0.01$, $w_{4}=10.0$, $\gamma^{control}_{max}=0.5$, $N^{control}_{batch}=10$, $N^{control}_{epoch}=10$とする.
}%

\subsection{State Estimator Using RMAE} \label{subsec:state-estimation}
\switchlanguage%
{%
  Since joint angles cannot usually be measured in musculoskeletal humanoids as described in \secref{sec:musculoskeletal-structure}, the joint angle of the actual robot is estimated by using vision.
  However, in this case, it is necessary to attach markers to the robot and for the robot to keep looking at self-body at all times.
  By estimating the current joint angle through RMAE from the information on current muscle tension and length, it is possible to keep tracking the condition of the self-body.
  The state estimation method is simple: the current muscle tension $\bm{f}^{cur}$ and the current muscle length $\bm{l}^{cur}$ are obtained from the actual robot, these values are inputted to RMAE in the form of (ii) in \figref{figure:network-structure}, and then the estimated joint angle $\bm{\theta}^{est}$ is obtained as $\bm{\theta}^{est} = \bm{h}_{dec, \bm{\theta}}(\bm{h}_{enc, \textrm{ii}}(\bm{f}^{cur}, \bm{l}^{cur}))$.

  Also, it is necessary for the state estimation method to be modified with the muscle rupture information $\bm{r}$.
  Since $\bm{f}$ and $\bm{l}$ of the ruptured muscle $i$, $f_i$ and $l_i$, cannot be trusted, the direct calculation of $\bm{z}$ from $\bm{h}_{enc, \textrm{ii}}$ results in incorrect values.
  There are two possible ways to solve this problem.

  The first method A is based on the assumption that there is no significant difference between the estimated joint angle in the previous frame $\bm{\theta}^{est, prev}$ and $\bm{\theta}^{cur}$.
  $\bm{f}^{cur}$ of the ruptured muscle $i$, $f^{cur}_{i}$, is set to zero, $\bm{l}^{pred}=\bm{h}_{dec, \bm{l}}(\bm{\theta}^{est, prev}, \bm{f}^{cur})$ is calculated, and $\bm{l}^{cur}$ of the ruptured muscle $i$, $l^{cur}_{i}$, is updated as $l^{cur}_{i}=l^{pred}_{i}$.
  Then, as usual, $\bm{\theta}^{est}$ is calculated as $\bm{\theta}^{est} = \bm{h}_{dec, \bm{\theta}}(\bm{h}_{enc, \textrm{ii}}(\bm{f}^{cur}, \bm{l}^{cur}))$.

  The second method A' is to update the latent variable $\bm{z}$ in the same way as in \secref{subsec:controller}.
  First, $\bm{z}=\bm{h}_{enc, \textrm{ii}}(\bm{f}^{cur}, \bm{l}^{cur})$ is calculated as with the unruptured muscle state.
  However, because $f_i$ and $l_i$ are unusual values, the same values of $\bm{f}$ and $\bm{l}$ cannot be reconstructed from $\bm{h}_{dec}(\bm{z})$.
  Therefore, the following loss function $L_{estimate}$ is calculated, and the current latent state $\bm{z}$ is obtained by updating $\bm{z}$ in the same way as in \secref{subsec:controller} (with the same parameters).
  \begin{align}
    L_{estimate}(\bm{f}, \bm{l}) = &w_{5}||f_i||_{2} + w_{6}||\bm{r}\otimes(\bm{f}-\bm{f}^{cur})||_{2}\nonumber\\ + &w_{7}||\bm{r}\otimes(\bm{l}-\bm{l}^{cur})||_{2}
  \end{align}
  Finally, the estimated joint angle is obtained as $\bm{\theta}^{est}=\bm{h}_{dec, \bm{\theta}}(\bm{z})$.

  Although it is clear that method A' is more accurate than method A, it is difficult to execute method A' with a fast cycle due to the high computational complexity of iterations.
  On the other hand, while method A is questionable in terms of the accuracy, it can be executed with a fast cycle.

  In this study, we set $w_{5}=10.0$, $w_{6}=1.0$, and $w_{7}=1.0$.
}%
{%
  状態推定の手法について述べる.
  筋骨格ヒューマノイドは\secref{sec:musculoskeletal-structure}で述べたように一般的には関節角度を測定することができないため, 視覚を用いて実機関節角度を推定する.
  しかしこの場合, 身体にマーカ等をつける必要や, 身体を常に見続けなければならない等の制約を受けることになる.
  そこで, 現在の筋張力・筋長情報からRMAEを通して関節角度を推定することで, 自身の状態を常に知り続けることが可能となる.

  方法は非常に単純で, 実機から現在筋張力$\bm{f}^{cur}$, 現在筋長$\bm{l}^{cur}$を取得し, これを\figref{figure:network-structure}の(ii)の形でネットワークに入力し, 関節角度推定値$\bm{\theta}^{est} = \bm{h}_{dec, \bm{\theta}}(\bm{h}_{enc, \textrm{ii}}(\bm{f}^{cur}, \bm{l}^{cur}))$を得るのみである.

  また, 筋が切れたという情報$\bm{r}$に対して, 状態推定を変化させる必要がある.
  これまでと同様, 切れた筋$i$の$f_i$, $l_i$が信用できないため, 直接$\bm{h}_{enc, \textrm{ii}}$により$\bm{z}$を求めてしまうと, おかしな値が出力されてしまう.
  これを解決する方法としては以下の2つが考えられる.

  一つ目は, 前フレームにおける推定関節角度$\bm{\theta}^{est, prev}$と現在の$\bm{\theta}^{cur}$には大きな違いが無いという仮定を使う方法である(手法A).
  切れた筋$i$の$\bm{f}^{cur}$である$f^{cur}_{i}=0$として, $\bm{l}^{pred}=\bm{h}_{dec, \bm{l}}(\bm{\theta}^{est, prev}, \bm{f}^{cur})$を計算し, $l^{cur}_{i}=l^{pred}_{i}$と更新する.
  その後, 通常と同様に$\bm{\theta}^{est} = \bm{h}_{dec, \bm{\theta}}(\bm{h}_{enc, \textrm{ii}}(\bm{f}^{cur}, \bm{l}^{cur}))$を計算するのみである.

  ニつ目は, \secref{subsec:controller}と同様の形で潜在変数$\bm{z}$を繰り返し更新していく方法である(手法A').
  まず筋が切れていない状態と同様に$\bm{z}=\bm{h}_{enc, \textrm{ii}}(\bm{f}^{cur}, \bm{l}^{cur})$を計算する.
  しかし, これは$f_i$, $l_i$が通常あり得ない値となっているため, $\bm{h}_{dec}(\bm{z})$により$\bm{f}$, $\bm{l}$を復元しても, 同じ値が出るわけではない.
  そこで, 以下の損失関数$L_{estimate}$を計算し, \secref{subsec:controller}と同様の形(パラメータも同じとする)で$\bm{z}$を更新していくことで, 現在状態を表す$\bm{z}$を得ることができる.
  \begin{align}
    L_{estimate}(\bm{f}, \bm{l}) = &w_{5}||f_i||_{2} + w_{6}||\bm{r}\otimes(\bm{f}-\bm{f}^{cur})||_{2}\nonumber\\ + &w_{7}||\bm{r}\otimes(\bm{l}-\bm{l}^{cur})||_{2}
  \end{align}
  最後に, $\bm{\theta}^{est}=\bm{h}_{dec, \bm{\theta}}(\bm{z})$により関節角度推定値を得る.

  手法A'の方が正確であることは明らかであるが, 繰り返しに時間がかかるため速い周期で実行することは難しい.
  それに対して手法Aは, 正確性には疑問が残るものの速い周期で実行することが可能である.

  本研究では, $w_{5}=10.0$, $w_{6}=1.0$, $w_{7}=1.0$とする.
}%

\subsection{Anomaly Detector Using RMAE} \label{subsec:detector}
\switchlanguage%
{%
  Data used in online learning is accumulated, and when the number of data exceeds $N^{detect}_{data}$, the model for anomaly detection is repeatedly rebuilt every time a new data is accumulated.
  At the same time, the data which exceeds $N^{detect}_{data}$ are discarded in order from the oldest data.
  For all the obtained data $(\bm{\theta}, \bm{f}, \bm{l})$, the predicted values $\{\bm{\theta}, \bm{f}, \bm{l}\}^{pred}$ are calculated as $(\bm{\theta}^{pred}, \bm{f}^{pred}, \bm{l}^{pred}) = \bm{h}_{dec}(\bm{h}_{enc, \textrm{ii}}(\bm{f}, \bm{l}))$.
   For each muscle $j$, $\bm{v}_{j}=\begin{pmatrix}f_j&l_j\end{pmatrix}^T$ and $\bm{v}^{pred}_{j}=\begin{pmatrix}f^{pred}_{j}&l^{pred}_{j}\end{pmatrix}^T$ are constructed.
  The average $\mu_{e}$ and covariance matrix $\Sigma_{e}$ for all data of $\bm{e}_{j}=\bm{v}_{j}-\bm{v}^{pred}_{j}$ are calculated.

  In the anomaly detection phase, $\bm{f}^{cur}$ and $\bm{l}^{cur}$ are always obtained, $\bm{f}^{pred}$ and $\bm{l}^{pred}$ are predicted, and the following value of Mahalanobis Distance is calculated for each muscle $j$.
  \begin{align}
    \bm{e}_{j}&=\begin{pmatrix}\bm{f}^{cur}_{j}&\bm{l}^{cur}_{j}\end{pmatrix}^{T}-\begin{pmatrix}\bm{f}^{pred}_{j}, \bm{l}^{pred}_{j}\end{pmatrix}^{T}\\
      d_j&=\sqrt{(\bm{e}_{j}-\mu_{e})^{T}\Sigma^{-1}_{e}(\bm{e}_{j}-\mu_{e})}
  \end{align}
  When $d_j$ exceeds a threshold $C^{detect}_{thre}$, it is considered that an anomaly has emerged.
  When an anomaly is detected, the anomaly detection and online update of RMAE are stopped.

  In this study, we set $N^{detect}_{data}=50$, and $C^{detect}_{thre}=30.0$.
  $C^{detect}_{thre}$ is determined from preliminary experiments, since no anomalies are detected when the value is too low and anomalies continue to be detected when the value is too high.
}%
{%
  オンライン学習の際のデータを蓄積していき, その個数が$N^{detect}_{data}$を超えたところから, データが貯まる度に異常検知を行うためのモデルを構築し直すことを繰り返す.
  同時に$N^{detect}_{data}$を超えたデータは古いものから捨てていく.
  得られた全データ$(\bm{\theta}, \bm{f}, \bm{l})$に対して, $(\bm{\theta}^{pred}, \bm{f}^{pred}, \bm{l}^{pred}) = \bm{h}_{dec}(\bm{h}_{enc, \textrm{ii}}(\bm{f}, \bm{l}))$のように, 予測値を計算する.
  それぞれの筋$j$に対して, $\bm{v}_{j}=(f_j, l_j)^T$, $\bm{v}^{pred}_{j}=(f^{pred}_{j}, l^{pred}_{j})^T$を作成する.
  この誤差$\bm{e}_{j}=\bm{v}_{j}-\bm{v}^{pred}_{j}$に関する全データの平均$\mu_{e}$と共分散行列$\Sigma_{e}$を計算する.

  実際の異常検知では, 現在の$\bm{f}^{cur}$, $\bm{l}^{cur}$を常に取得し, 同様に$\bm{f}^{pred}$, $\bm{l}^{pred}$を予測して, 以下の値をそれぞれの筋$j$に対して計算する.
  \begin{align}
    \bm{e}_{j}&=\begin{pmatrix}\bm{f}^{cur}_{j}&\bm{l}^{cur}_{j}\end{pmatrix}^{T}-\begin{pmatrix}\bm{f}^{pred}_{j}, \bm{l}^{pred}_{j}\end{pmatrix}^{T}\\
      d_j&=\sqrt{(\bm{e}_{j}-\mu_{e})^{T}\Sigma^{-1}_{e}(\bm{e}_{j}-\mu_{e})}
  \end{align}
  この$d_j$の値が閾値$C^{detect}_{thre}$を超えた時, 異常が検知されたと見なす.
  異常が検知された際は, 一旦異常検知やRMAEのモデルの更新を停止する.

  本研究では$N^{detect}_{data}=50$, $C^{detect}_{thre}=30.0$とした.
  $C^{detect}_{thre}$は低すぎると検知されなくなり, 高すぎると異常が検知され続けてしまうため, 予備実験により値を決定している.
}%


\begin{figure}[t]
  \centering
  \includegraphics[width=0.8\columnwidth]{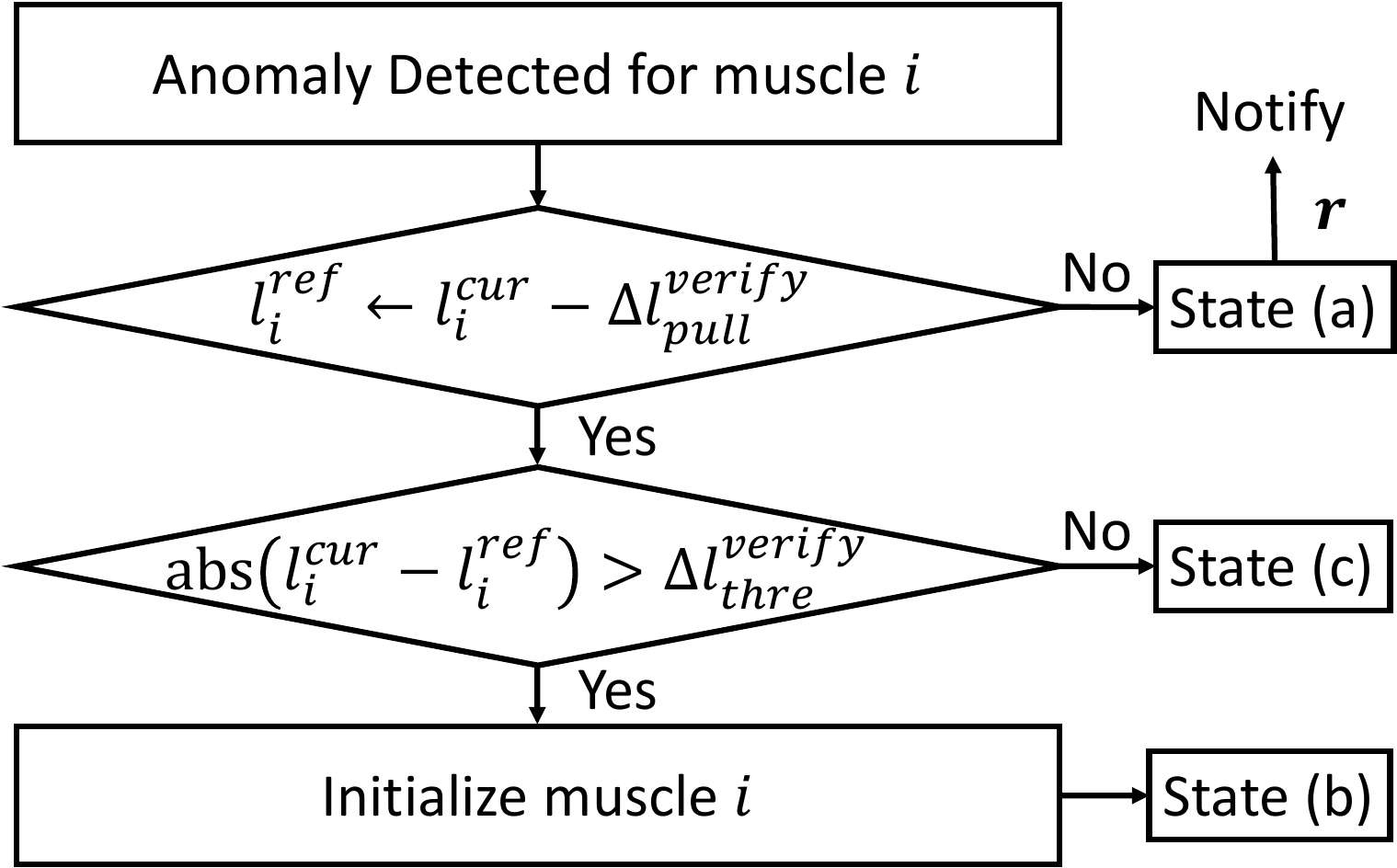}
  \caption{The procedure of muscle rupture verification.}
  \label{figure:verification}
\end{figure}

\subsection{Muscle Rupture Verification} \label{subsec:verification}
\switchlanguage%
{%
  When an anomaly is detected, it is necessary to verify the current state of the muscle rupture.
  In this study, we can consider the following states of muscles: (a) the muscle has broken or cannot transmit force due to failure of the motor, motor driver, etc., (b) the muscle has broken but can continue to be used depending on the broken part of the muscle wire, and (c) the muscle is not broken even though an anomaly is detected.
  The state (b) can occur when the end point of the muscle falls off of a fixed point or when the wire between the end point of the muscle and a nonlinear elastic element breaks.
  In this case, the nonlinear elastic element is trapped in a muscle relay unit that folds back the muscle, and although the origin of the muscle length is offset, it can still be used.
  The procedure for determining the muscle state is shown in \figref{figure:verification}.

  The control of \equref{eq:muscle-stiffness-control} ensures that the muscle is not loosened.
  In order to prevent overwinding of the muscle after the muscle rupture, $\bm{l}^{cur}$ is restricted to wind only $\Delta{l}^{verify}_{max}$ from $\bm{l}^{ref}$ at maximum.
  The following procedure is performed for all muscles in which an anomaly is detected.
  First, if the muscle $i$ is not ruptured and is working properly, the muscle tension will increase when $l^{ref}_{i}$ is set to $l^{cur}_{i}-\Delta{l}^{verify}_{pull}$.
  In other words, if the change in muscle tension $\Delta{f}_{i}$ exceeds $\Delta{f}^{verify}_{thre}$, it is considered that the muscle is working properly: state (b) or (c).
  If the muscle is not working properly, the muscle is assumed to be ruptured: state (a).
  Then, the muscle rupture information $\bm{r}$ with the value of muscle $i$ as zero is notified to all components, and these execution methods are partially changed.
  If the muscle is working properly, we focus on the magnitude of $\textrm{abs}(l^{cur}_{i}-l^{ref}_{i})$.
  If this value is greater than $\Delta{l}^{verify}_{thre}$, the muscle state is considered to be (b), and if it is less than $\Delta{l}^{verify}_{thre}$, the muscle state is considered to be (c).

  In the case of (b), although force is transmitted, the origin of the muscle length is offset and needs to be initialized.
  Here again, the same method as in \secref{subsec:controller} is used, where the latent variable $\bm{z}$ is repeatedly updated.
  First, the initial $\bm{z}$ is calculated as $\bm{z}=\bm{h}_{enc, \textrm{i}}(\bm{\theta}^{cur}, \bm{f}^{cur})$.
  Then, since values other than $l^{cur}_{i}$ are correct, $\bm{z}$ representing the current state can be obtained by calculating the following loss function $L_{verify}$ considering the match to the current joint angle, muscle tension, and muscle length, and by updating $\bm{z}$ in the same way as in \secref{subsec:controller} (with the same parameters),
  \begin{align}
    L_{verify}(\bm{\theta}, \bm{f}, \bm{l}) = &w_{8}||\bm{\theta}-\bm{\theta}^{cur}||_{2} + w_{9}||\bm{f}-\bm{f}^{cur}||_{2}\nonumber\\ + &w_{10}||\bm{r}^{offset}\otimes(\bm{l}-\bm{l}^{cur})||_{2}
  \end{align}
  where $\bm{r}^{offset}$ is an $M$-dimensional vector where only the offset muscle $i$ is 0 and the rest are 1.
  Finally, the estimated muscle length is obtained from $\bm{l}^{est}=\bm{h}_{dec, \bm{l}}(\bm{z})$, and $l^{cur}_{i}$ is initialized by $l^{est}_{i}$.

  Since the body structure of the robot changes in the case of (a) and (b), the model of RMAE is stored in a different place and is restored after the muscle module is fixed or replaced.
  In addition, in the case of (a) and (b), all the accumulated data for constructing RMAE and the anomaly detection model are deleted.
  Once this procedure is finished for all the anomaly-detected muscles, the online learning of RMAE and updating of the anomaly detection model is resumed as usual.

  In this study, we set $\Delta{l}^{verify}_{max}=100$ [mm], $\Delta{l}^{verify}_{pull}=10$ [mm], $\Delta{f}^{verify}_{thre}=10$ [N], $\Delta{l}^{verify}_{thre}=30$ [mm], $w_{8}=1.0$, $w_{9}=1.0$, and $w_{10}=1.0$.
}%
{%
  異常が検知されたとき, 現在の筋状態がどうなっているのかを理解する必要がある.
  本研究では筋の状態に, (a)筋が切れた, またはモータの故障等によって力を伝達することができない状態, (b)筋は切れたが切れる部分によっては継続して使える状態, (c)異常が検知されたものの筋は切れていなかった状態, を考える.
  (b)については, 筋の端点が固定点から取れる, または筋の端点から非線形性要素までの間のワイヤが切れることによって起こりうる.
  この場合, 筋を折り返す経由点ユニット等に外れた非線形性要素が引っ掛かり, 筋の0点はオフセットしてしまうものの, まだ利用することが可能である.
  これらを判断する手順を\figref{figure:verification}に示す.

  まず, 筋は\equref{eq:muscle-stiffness-control}の制御によって, 緩まない状態になっている.
  なお, 筋が切れた場合に巻き過ぎるのを防止するため$\bm{l}^{cur}$は$\bm{l}^{ref}$よりも$\Delta{l}^{verify}_{max}$以上は巻かないという制限を加えている.
  以降の操作を異常が検知された全ての筋に対して実行する.
  まず異常が検知された筋$i$について, $l^{ref}_{i}$を$l^{cur}_{i}$から$\Delta{l}^{verify}_{pull}$だけ巻いた位置に設定した時に, もし筋が切れておらず正しく働いていれば, 筋張力が上がることになる.
  つまり, 筋張力の変化$\Delta{f}_{i}$が$\Delta{f}^{verify}_{thre}$を超えた時, 筋は正常に働いているとする.
  もし筋が正常に働いていない場合は, その筋は切れていると見なし, その筋$i$の値を0とした$\bm{r}$を全コンポーネントに通達し, それぞれの実行の仕方を変化させる.
  もし筋が正常に働いている場合は, $\textrm{abs}(l^{cur}_{i}-l^{ref}_{i})$の大きさに着目する.
  この値が$\Delta{l}^{verify}_{thre}$よりも大きかった場合は(b)の状態であり, それよりも小さかった場合は(c)であると見なす.

  (b)の場合は, 力は伝達するものの, 筋長の原点がオフセットしてしまっているため, これを初期化する必要がある.
  ここでも, \secref{subsec:controller}と同様の形で潜在変数$\bm{z}$を繰り返し更新していく方法を用いる.
  まず$\bm{z}=\bm{h}_{enc, \textrm{i}}(\bm{\theta}^{cur}, \bm{f}^{cur})$を計算する.
  そして, $l^{cur}_{i}$以外は正しい値のため以下の損失関数$L_{verify}$を計算し, \secref{subsec:controller}と同様の形(パラメータも同じとする)で$\bm{z}$を更新していくことで, 現在状態を表す$\bm{z}$を得ることができる.
  \begin{align}
    L_{verify}(\bm{\theta}, \bm{f}, \bm{l}) = &w_{8}||\bm{\theta}-\bm{\theta}^{cur}||_{2} + w_{9}||\bm{f}-\bm{f}^{cur}||_{2}\nonumber\\ + &w_{10}||\bm{r}^{offset}\otimes(\bm{l}-\bm{l}^{cur})||_{2}
  \end{align}
  ここで, $\bm{r}^{offset}$はオフセットしてしまった筋$i$のみ0で残りは1となる$M$次元のベクトルとする.
  最後に, $\bm{l}^{est}=\bm{h}_{dec, \bm{l}}(\bm{z})$により筋長推定値を得て, $l^{cur}_{i}$を$l^{est}_{i}$により初期化する.

  また, (a)と(b)の場合はロボットの身体構造が変化してしまうため, それまでに学習されたRMAEのモデルは別の場所に保存しておき, 筋が換装された後にそのモデルを復帰させる.
  加えて, (a)と(b)の場合はRMAE・異常検知モデルの学習・構築のための蓄積されたデータは全て削除する.
  異常検知された全ての筋についてこの操作が終わったら, 通常通りオンライン学習・異常検知モデルの更新等を再開する.

  本研究では, $\Delta{l}^{verify}_{max}=100$ [mm], $\Delta{l}^{verify}_{pull}=10$ [mm], $\Delta{f}^{verify}_{thre}=10$ [N], $\Delta{l}^{verify}_{thre}=30$ [mm], $w_{8}=1.0$, $w_{9}=1.0$, $w_{10}=1.0$とする.
}%

\begin{figure}[t]
  \centering
  \includegraphics[width=0.5\columnwidth]{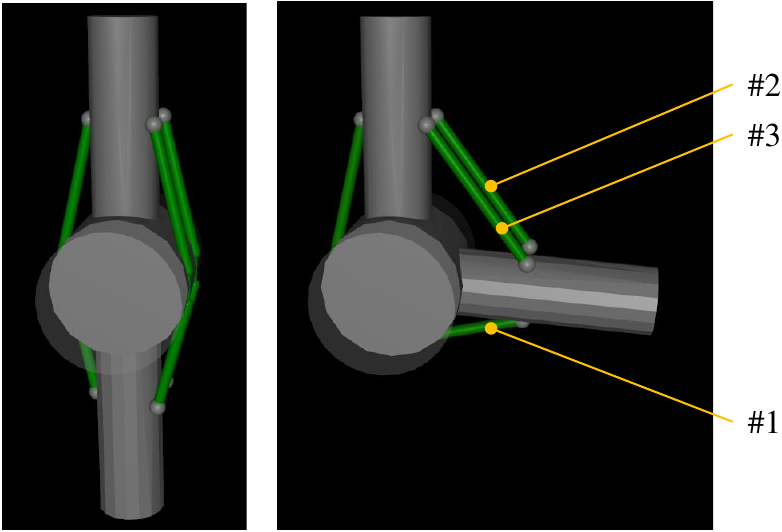}
  \caption{1 DOF simple joint model in MuJoCo \cite{todorov2012mujoco}.}
  \label{figure:simulation-setup}
\end{figure}

\begin{figure}[t]
  \centering
  \includegraphics[width=0.99\columnwidth]{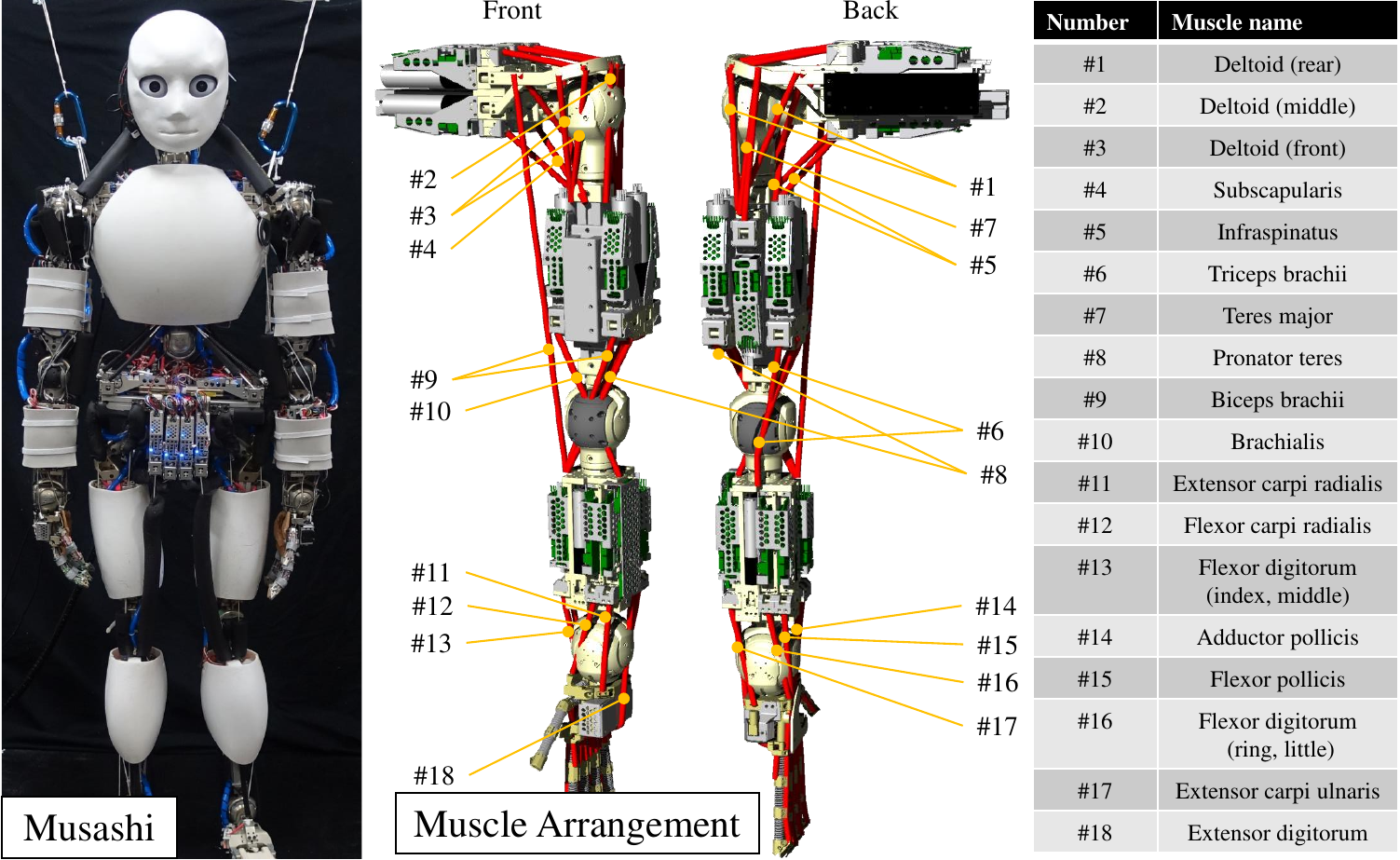}
  \caption{The musculoskeletal humanoid Musashi \cite{kawaharazuka2019musashi} and its muscle arrangement.}
  \label{figure:musashi-setup}
\end{figure}

\section{Experiments} \label{sec:experiments}
\subsection{Experimental Setup}
\switchlanguage%
{%
  First, a simple 1 DOF simulation is constructed using MuJoCo \cite{todorov2012mujoco} (\figref{figure:simulation-setup}).
  The structure of the elbow is imitated, and 3 redundant muscles are arranged.
  Nonlinear elastic elements are arranged and friction loss is set in order to be closer to the actual robot.
  While the geometric model of the muscle route is based on straight wire lines, the muscle route in simulation wraps around a cylinder of the elbow joint, and therefore requires learning of RMAE, just as in the actual robot.

  Second, the musculoskeletal humanoid Musashi \cite{kawaharazuka2019musashi} is used (\figref{figure:musashi-setup}).
  There are 74 muscles in the body, and nonlinear elastic elements are placed at the ends of the muscles.
  Also, although ordinary musculoskeletal humanoids cannot measure joint angle, Musashi is equipped with a special mechanism called joint module, which can measure joint angle for experimental evaluation and can also be used as $\bm{\theta}^{est'}$.
  Among the joints of Musashi, we mainly use 5 DOFs of the arm, denoted as S-p, S-r, S-y, E-p, and E-y (S and E expresses the shoulder and elbow, and rpy expresses roll, pitch and yaw, respectively).
  The number of muscles related to these 5 DOFs is 10, including 1 polyarticular muscle.

  In this study, the effects of muscle rupture on each component and the behavior of changes in the components after muscle rupture are tested in both simulation and actual robot.
  The effectiveness of this study is demonstrated through a series of motion experiments with these components utilizing the redundant muscle arrangement.
  Note that for state estimation, the difference between the estimated joint angle $\bm{\theta}^{est}$ obtained by \secref{subsec:state-estimation} and the current joint angle of the robot $\bm{\theta}^{cur}$, $\textrm{RMSE}_{est}=||\bm{\theta}^{est}-\bm{\theta}^{cur}||_{2}$, is compared, and for control, the difference between the target joint angle $\bm{\theta}^{ref}$ and $\bm{\theta}^{cur}$, $\textrm{RMSE}_{control}=||\bm{\theta}^{ ref}-\bm{\theta}^{cur}||_{2}$, is compared.
  All the parameters set in this study are listed at the end of each subsection in Sec. 3.
}%
{%
  まず, 1自由度のシミュレーションをMuJoCo \cite{todorov2012mujoco}で作成した(\figref{figure:simulation-setup}).
  肘の構造を模しており, 筋が3本と冗長に配置されている.
  実機に近づくよう, 非線形弾性要素を配置しており, また, 筋に摩擦損失を持たせている.
  筋経路の幾何モデルは経由点を直線で結んだものであるのに対して, シミュレーションでは円筒に巻き付く形をしているため, 実機同様RMAEの学習が必要な状況を作り出している.

  次に, 筋骨格ヒューマノイドMusashi \cite{kawaharazuka2019musashi}を用いる(\figref{figure:musashi-setup}).
  全身の筋は74本であり, 筋の末端には非線形弾性要素が配置されている.
  また, 通常の筋骨格ヒューマノイドでは関節角度は測定出来ないが, Musashiには関節モジュールという特殊な機構により, 実験評価のための関節角度センサが配置されており, $\bm{\theta}^{est'}$としても用いることができる.
  その中でも, 主に使用するのは腕の5つの関節自由度であり, これらをS-p, S-r, S-y, E-p, E-yとする(S, Eはそれぞれ肩と肘を表し, rpyはそれぞれroll, pitch, yawを表す).
  これら5自由度に関係する筋は10本であり, 2関節筋を一つ含む.

  本研究では, 筋破断時の各コンポーネントへの影響, また, これまで説明したコンポーネントの変化による筋破断への対処法について, シミュレーション・実機において順に実験を行う.
  また, これらコンポーネントを用いた, 冗長な筋配置を活かした一連の動作実験を行い, 本研究の有効性を示す.
  なお, 状態推定については\secref{subsec:state-estimation}で得られる推定関節角度$\bm{\theta}^{est}$と現在のロボットの関節角度$\bm{\theta}^{cur}$の差である$\textrm{RMSE}_{est}=||\bm{\theta}^{est}-\bm{\theta}^{cur}||_{2}$を, 制御については, 指令関節角度$\bm{\theta}^{ref}$と$\bm{\theta}^{cur}$の差である$\textrm{RMSE}_{control}=||\bm{\theta}^{ref}-\bm{\theta}^{cur}||_{2}$を測定し比較する.

}%

\subsection{1 DOF Simulation Experiment}
\switchlanguage%
{%
  The online learning, state estimation, and control experiments are conducted using a simple 1 DOF joint simulator.
  In this section, $\bm{\theta}$ is a 1-dimensional vector, and $\{\bm{l}, \bm{f}\}$ is a 3-dimensional vector.
}%
{%
  1自由度のシミュレーションを用いて, 筋長制御・状態推定・オンライン学習について実験を行う.
  本節では, $\bm{\theta}$は1自由度, $\{\bm{l}, \bm{f}\}$は3自由度である.
}%

\begin{figure}[t]
  \centering
  \includegraphics[width=0.99\columnwidth]{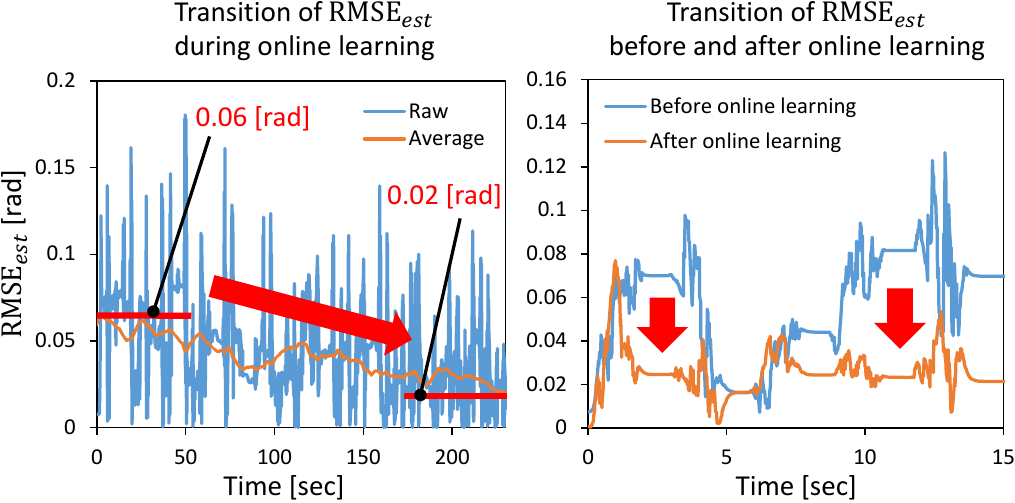}
  \caption{1 DOF simulation experiment of state estimation without any muscle ruptured: transition of $\textrm{RMSE}_{est}$ during online learning (left graph), and comparison of $\textrm{RMSE}_{est}$ between before and after online learning (right graph).}
  \label{figure:sim-state-learning}
\end{figure}

%

\begin{figure*}[t]
  \centering
  \includegraphics[width=1.99\columnwidth]{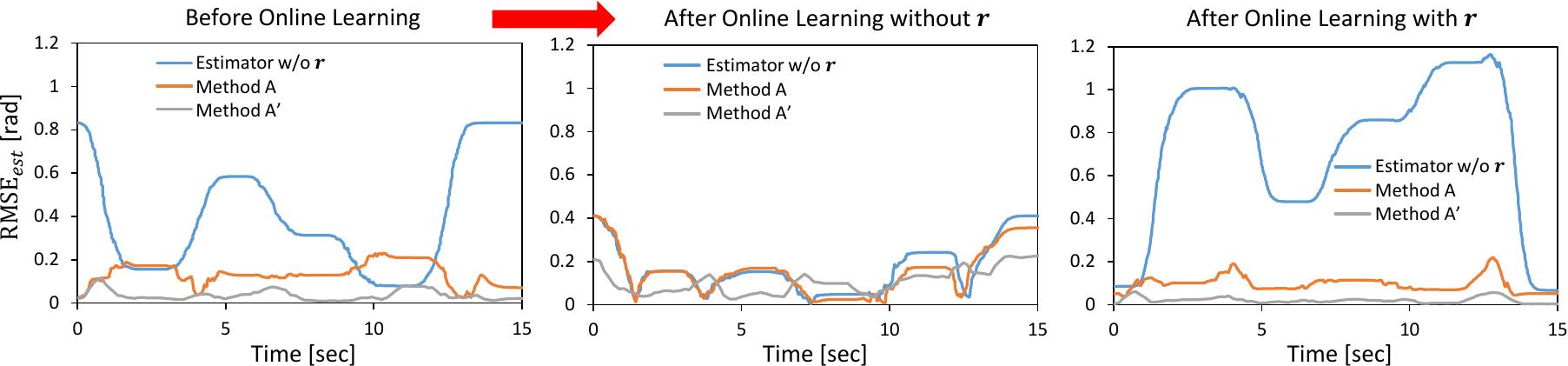}
  \vspace{-1.0mm}
  \caption{1 DOF simulation experiment of state estimation when muscle \#2 is ruptured: comparison of transitions of $\textrm{RMSE}_{est}$ among without $\bm{r}$, with method A, and with method A', before online learning (left graph), after online learning without $\bm{r}$ (center graph), and after online learning with $\bm{r}$ (right graph).}
  \label{figure:sim-state-all}
  \vspace{-1.0mm}
\end{figure*}

\subsubsection{Online Learning and State Estimator Experiment} \label{subsubsec:sim-state-estimation}
\switchlanguage%
{%
  First, we repeatedly send random joint angles $\bm{\theta}^{rand}$ within the joint angle range, while executing the online learning of \secref{subsec:online-learning}.
  Here, various states can be obtained as training data by not only using the control method of \secref{subsec:controller} but also sending the muscle length obtained from $\bm{h}_{dec, \bm{l}}(\bm{h}_{enc, \textrm{i}}(\bm{\theta}^{rand}, \bm{f}^{rand}))$ ($\bm{f}^{rand}$ is a random muscle tension).
  For each random joint angle, in turn, the muscle length is sent over 2 seconds by the method of \secref{subsec:controller}, rests for 0.5 seconds, is sent over 1 second by the above method, and rests for another 0.5 seconds.
  This procedure is repeated.
  The left graph of \figref{figure:sim-state-learning} shows the transition of $\textrm{RMSE}_{est}$ and its average of every 20 seconds.
  The average of $\textrm{RMSE}_{est}$ decreased gradually from about 0.06 to 0.02 rad in 4 minutes.
  Then, five random joint angles $\bm{\theta}^{rand}$ are specified, and in turn, muscle length is sent by the method of \secref{subsec:controller} over 2 seconds and rests for 0.5 seconds (we call this an evaluation experiment).
  The right graph of \figref{figure:sim-state-learning} shows how the transition of $\textrm{RMSE}_{est}$ changes when using RMAE before and after online learning.
  From this experiment onwards, online learning is not executed during the evaluation experiment.
  Most of the time, we found that $\textrm{RMSE}_{est}$ was smaller after the online learning than before.
  The results of the state estimator in 1 DOF simulation including the subsequent experiments are shown in \tabref{table:sim-state-compare}.
  The average of $\textrm{RMSE}_{est}$ during 15 seconds of the evaluation experiment, $\textrm{RMSE}^{ave}_{est}$, was 0.056 rad before online learning and 0.026 rad after online learning.

  Next, the left graph of \figref{figure:sim-state-all} shows the transition of $\textrm{RMSE}_{est}$ when muscle \#2 breaks and its muscle length is constant at 100 mm in the evaluation experiment.
  RMAE after the online learning as above is used.
  Here, we compare the three state estimations of \secref{subsec:state-estimation}, for without the muscle rupture information $\bm{r}$ and for two methods with $\bm{r}$ (A and A').
  Method A' had the smallest error in 15 seconds, followed by method A and without $\bm{r}$.
  $\textrm{RMSE}^{ave}_{est}$ was 0.40 rad without $\bm{r}$, 0.13 rad for method A, and 0.037 rad for method A'.
  In all cases, the error was larger than 0.026 rad after the online learning of RMAE, and we can see that unlearned states were created due to the muscle rupture, which increased the error of state estimation.

  Finally, the center and right graphs of \figref{figure:sim-state-all} show the transitions of $\textrm{RMSE}_{est}$ after online learning with muscle \#2 ruptured in the evaluation experiment.
  Here, we compare the online learnings in \secref{subsec:online-learning} for the cases without $\bm{r}$ (the center graph of \figref{figure:sim-state-all}) and with $\bm{r}$ (the right graph of \figref{figure:sim-state-all}).
  In each case, we compare the state estimations of \secref{subsec:state-estimation} for without $\bm{r}$ and for the two methods with $\bm{r}$ (method A and A').
  After the online learning without $\bm{r}$, there was no significant difference of the transitions among three methods (without $\bm{r}$, method A, and method A'), though the error of method A' was slightly smaller than that of others.
  $\textrm{RMSE}^{ave}_{est}$ was 0.17 rad for without $\bm{r}$, 0.15 rad for method A, and 0.11 rad for method A'.
  Therefore, we found that the results after online learning without $\bm{r}$ were worse than before, regarding method A and A'.
  On the other hand, the joint angle errors after online learning with $\bm{r}$ were small in the order of method A', method A, and without $\bm{r}$.
  $\textrm{RMSE}^{ave}_{est}$ was 0.73 rad without $\bm{r}$, 0.097 rad for method A, and 0.021 rad for method A'.
  Therefore, we can see that the error of state estimation is reduced by continuing online learning with $\bm{r}$, even if the muscle is ruptured.
  Note that in these experiments, method A can be executed at 100 Hz or higher, while method A' can only be executed at 10 Hz at most.
}%
{%
  まず, \secref{subsec:online-learning}のオンライン学習を実行した状態で, 関節角度範囲内のランダムな関節角度$\bm{\theta}^{rand}$を送ることを繰り返す.
  このとき, \secref{subsec:controller}の方法だけでなく, $\bm{h}_{dec, \bm{l}}(\bm{h}_{enc, \textrm{i}}(\bm{\theta}^{rand}, \bm{f}^{rand}))$により得られた筋長を送ることで様々な状態を学習データとして得ることができる($\bm{f}^{rand}$はランダムな筋張力).
  順に, \secref{subsec:controller}の方法で筋長を2秒で送り0.5秒休み, 上記の方法で筋長を1秒で送り0.5秒休むことを繰り返す.
  $\textrm{RMSE}_{est}$の遷移を\figref{figure:sim-state-learning}の左図に示す.
  また, 20秒ごとの平均の値も表示している.
  $\textrm{RMSE}_{est}$の平均は徐々に減っていき, 4分間かけて約0.06から0.02 radまで落ちていることがわかる.
  次に, ランダムな5つの関節角度$\bm{\theta}^{rand}$を指定し, これを順に\secref{subsec:controller}の方法で2秒で送って0.5秒休む動作を行った.
  このとき, オンライン学習前と後のRMAEを使った際, どの程度$\textrm{RMSE}_{est}$の遷移が変化するかを\figref{figure:sim-state-learning}の右図に示す.
  なお, 本実験からは評価実験中にオンライン学習は止めている.
  ほとんどの時間に置いて, 学習後は学習前に比べて$\textrm{RMSE}_{est}$が小さくなっていることがわかる.
  以降の実験を含めた1DOFシミュレーションにおける状態推定の結果を\tabref{table:sim-state-compare}に示す.
  15秒間の本動作における$\textrm{RMSE}_{est}$の平均, $\textrm{RMSE}^{ave}_{est}$は, 学習前は0.056 rad, 学習後は0.026 radであった.

  次に, オンライン学習後のRMAEを使って状態推定を行う際, 筋\#2が切れ, 筋長が100 mmで一定となってしまった場合に, 先と同じ実験を行ったときの$\textrm{RMSE}_{est}$の遷移を\figref{figure:sim-state-all}の左図に示す.
  ここで, \secref{subsec:state-estimation}における, 筋の破断情報$\bm{r}$無かった場合, あった場合の二つの手法(AとA')の場合における状態推定について比較を行う.
  15秒間を通して手法A'が最も誤差が少なく, 一部逆転する箇所はあるものの, ついで手法A, 情報$\bm{r}$が無い場合の順で誤差が小さい.
  $\textrm{RMSE}^{ave}_{est}$は, $\bm{r}$がない場合は0.40 rad, 手法Aは0.13 rad, 手法A'は0.037 radであった.
  どの状態についても, 学習後の0.026 radよりは誤差が大きく, 筋の破断により学習されていないような状態が生まれ誤差が増大していることがわかる.

  最後に, その筋が切れた状態でオンライン学習を行った後に, 先と同じ実験を行ったときの$\textrm{RMSE}_{est}$の遷移を\figref{figure:sim-state-all}の中図・右図に示す.
  ここで, \secref{subsec:online-learning}における, 筋の破断情報$\bm{r}$が無かった場合(\figref{figure:sim-state-all}の中図)とあった場合(\figref{figure:sim-state-compare}の右図)のオンライン学習について比較を行う.
  それぞれの場合について, \secref{subsec:state-estimation}については筋の破断情報$\bm{r}$がない場合, あった場合の二つの手法(手法AとA')の場合における状態推定について比較を行う.
  破断情報$\bm{r}$がない状態でオンライン学習を行った場合は, 状態推定において$\bm{r}$がない場合, 手法AまたはA'を使った場合について, A'の方が多少誤差が小さいものの, 誤差の遷移に大きな差はない.
  実際$\textrm{RMSE}^{ave}_{est}$は, $\bm{r}$がない場合は0.17 rad, 手法Aは0.15 rad, 手法A'は0.11 radであった.
  よって, 破断情報$\bm{r}$がない状態でオンライン学習を続けると, 手法A, A'については学習前よりも悪い結果となってしまうことがわかった.
  これに対して, 破断情報$\bm{r}$がある状態でオンライン学習を行った場合について, その誤差は順に, 手法A', 手法A, $\bm{r}$がない場合の順で小さい.
  $\textrm{RMSE}^{ave}_{est}$は, $\bm{r}$がない場合は0.73 rad, 手法Aは0.097 rad, 手法A'は0.021 radであった.
  よって, 筋が切れても$\bm{r}$の情報を含めたオンライン学習を続けることで, 状態推定の誤差が小さくなることがわかった.
  なお, これらの実験において, 手法Aは100 Hz以上で実行できるのに対して, 手法A'は最大で10 Hz程度までしか出ない.
}%

\begin{table}[htb]
  \centering
  \caption{Results of evaluation experiments for state estimator in 1 DOF simulation: the averages of $\textrm{RMSE}_{est}$ during 15 seconds of the evaluation experiment, $\textrm{RMSE}^{ave}_{est}$.}
  \vspace{0.1in}
  \scalebox{0.7}{
    \begin{tabular}{|l|c|c|c|c|c|} \hline
      Muscle state & \multicolumn{2}{c|}{w/o rupture} & \multicolumn{3}{c|}{w/ \#2 ruptured} \\ \hline
      Online learning & Before & After & Before & After, w/o $\bm{r}$ & After, w/ $\bm{r}$ \\ \hline
      Estimator w/o $\bm{r}$ [rad] & 0.056 & \textbf{0.026} & 0.40 & 0.17 & 0.73 \\
      Method A [rad] & - & - & 0.13 & 0.15 & 0.097 \\
      Method A' [rad] & - & - & 0.037 & 0.11 & \textbf{0.021} \\ \hline
    \end{tabular}
  }
  \label{table:sim-state-compare}
\end{table}

%
%

\begin{figure*}[t]
  \centering
  \includegraphics[width=1.99\columnwidth]{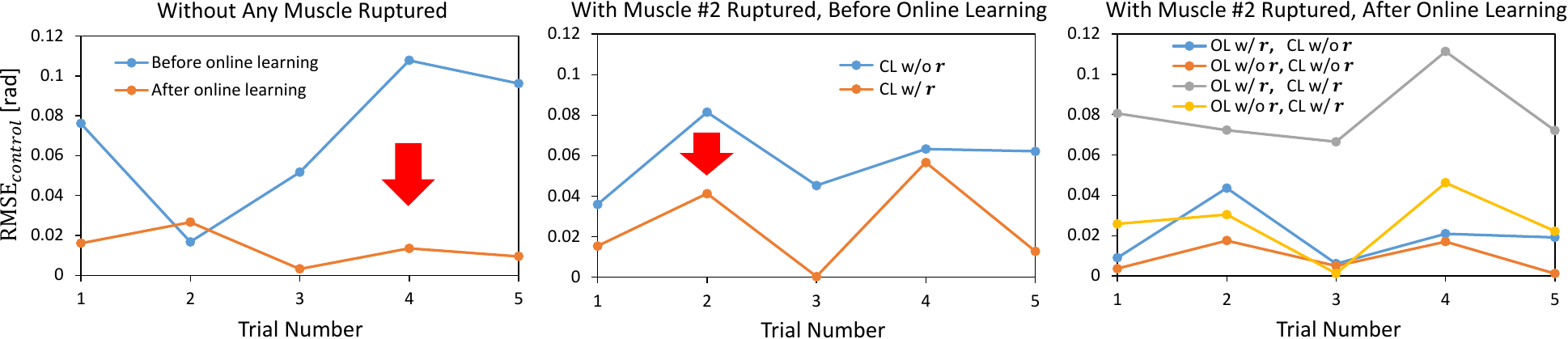}
  \vspace{-1.0mm}
  \caption{1 DOF simulation experiment of controller: comparison of transitions of $\textrm{RMSE}_{control}$ between before and after online learning without any muscle ruptured (left graph), between using controller with or without $\bm{r}$ with muscle \#2 ruptured (center graph), and between after online learning with or without $\bm{r}$ and using controller with or without $\bm{r}$ with muscle \#2 ruptured (right graph). OL means online learning and CL means controller.}
  \label{figure:sim-control-all}
  \vspace{-1.0mm}
\end{figure*}

\subsubsection{Online Learning and Controller Experiment} \label{subsubsec:sim-control}
\switchlanguage%
{%
  We execute the evaluation experiment as in \secref{subsubsec:sim-state-estimation} to see the control error.
  RMAEs before and after online learning in \secref{subsubsec:sim-state-estimation} are used for the controller, and $\textrm{RMSE}_{control}$ is calculated after each motion at rest.
  The results are shown in the left figure of \figref{figure:sim-control-all}.
  Although the magnitude of the error at the second joint angle was slightly reversed, we can see that the target joint angle was generally realized more accurately after online learning than before.
  The results of the controller in 1 DOF simulation including the subsequent experiments are shown in \tabref{table:sim-control-compare}.
  The average of $\textrm{RMSE}_{control}$ for the five joint angles, $\textrm{RMSE}^{ave}_{control}$, was 0.070 rad before online learning and 0.014 rad after online learning.

  Next, we show the transition of $\textrm{RMSE}_{control}$ when muscle \#2 is ruptured as in \secref{subsubsec:sim-state-estimation}, in the center graph of \figref{figure:sim-control-all}.
  Here, we compare the controllers in \secref{subsec:controller} with and without $\bm{r}$.
  We can see that the error of the controller with $\bm{r}$ was smaller than that without $\bm{r}$ regarding all joint angles.
  $\textrm{RMSE}^{ave}_{control}$ was 0.058 rad without $\bm{r}$ and 0.025 rad with $\bm{r}$.
  Also, we can see that the error of joint angle after muscle rupture was larger than that before muscle rupture, regardless of with or without $\bm{r}$.

  Finally, we show the results of the evaluation experiment after online learning with the muscle ruptured, in the right figure of \figref{figure:sim-control-all}.
  Here, we compare the online learnings of \secref{subsec:online-learning} with and without $\bm{r}$.
  For each case, we also compare the controllers of \secref{subsec:controller} with and without $\bm{r}$.
  The results show that the error was smaller when $\bm{r}$ was available for the controller, in each case after online learning with or without $\bm{r}$.
  Similarly, for the online learning, we can see that the error was smaller when $\bm{r}$ was available.
}%
{%
  前節と同様にランダムな5つの関節角度$\bm{\theta}^{rand}$を$\bm{\theta}^{ref}$に指定し, これを順に\secref{subsec:controller}の方法で2秒で送って0.5秒休む動作を行う.
  このとき, 前節のオンライン学習前と学習後のRMAEを使い, それぞれの動作後の静止時における$\textrm{RMSE}_{control}$を計算する.
  その結果を\figref{figure:sim-control-all}の左図に示す.
  2つめの関節角度では多少逆転しているものの, おおむね学習後は学習前に比べて正確な関節角度を実現できていることがわかる.
  以降の実験を含めた1DOFシミュレーションにおける制御の結果を\tabref{table:sim-control-compare}に示す.
  5回の本動作における$\textrm{RMSE}_{control}$の平均$\textrm{RMSE}^{ave}_{control}$は, 学習前は0.070 rad, 学習後は0.014 radであった.

  次に, 前節と同様に筋\#2が切れてしまった場合について, 先と同じ実験を行ったときの$\textrm{RMSE}_{control}$を\figref{figure:sim-control-all}の中図に示す.
  ここで, \secref{subsec:controller}における, 筋の破断情報$\bm{r}$が無い場合とある場合の筋長制御について比較を行う.
  グラフから, どの状態においても情報$\bm{r}$があった場合の方が, 無かった場合に比べて誤差が小さいことがわかる.
  $\textrm{RMSE}^{ave}_{control}$は, $\bm{r}$がない場合は0.058 rad, $\bm{r}$がある場合は0.025 radであった.
  よって, 筋が切れた際, $\bm{r}$の情報の有無に関わらず, 学習後よりも誤差が大きくなっていることもわかる.

  最後に, その筋が切れた状態でオンライン学習を行った後に, 先と同じ実験を行ったときの$\textrm{RMSE}_{control}$を\figref{figure:sim-control-all}の右図に示す.
  ここで, \secref{subsec:online-learning}における, 筋の破断情報$\bm{r}$がない場合とある場合のオンライン学習について比較を行う.
  また, それぞれについて, \secref{subsec:controller}における筋の破断情報$\bm{r}$がない場合とある場合の筋長制御に関する比較も行う.
  $\bm{r}$がない場合とある場合のオンライン学習についてそれぞれ, 筋長制御には$\bm{r}$の情報があった方が誤差が少ないことがわかる.
  また, オンライン学習についても同様に, $\bm{r}$の情報があった方が誤差が少ないことがわかる.
}%

\begin{table}[htb]
  \centering
  \caption{Results of evaluation experiments for controller in 1 DOF simulation: the averages of $\textrm{RMSE}_{control}$ for the five joint angles of the evaluation experiment, $\textrm{RMSE}^{ave}_{control}$.}
  \vspace{0.1in}
  \scalebox{0.7}{
    \begin{tabular}{|l|c|c|c|c|c|} \hline
      Muscle state & \multicolumn{2}{c|}{w/o rupture} & \multicolumn{3}{c|}{w/ \#2 ruptured} \\ \hline
      Online learning & Before & After & Before & After, w/o $\bm{r}$ & After, w/ $\bm{r}$ \\ \hline
      Controller w/o $\bm{r}$ [rad] & 0.070 & \textbf{0.014} & 0.058 & 0.081 & 0.020 \\
      Controller w/ $\bm{r}$ [rad] & - & - & 0.025 & 0.025 & \textbf{0.0089} \\\hline
    \end{tabular}
  }
  \label{table:sim-control-compare}
\end{table}

\subsection{Actual Robot Experiment}
\switchlanguage%
{%
  The online learning, state estimation, control, anomaly detection, and muscle rupture verification experiments are conducted using the actual robot of the left arm of Musashi.
  In this section, $\bm{\theta}$ is a 5-dimensional vector, and $\{\bm{l}, \bm{f}\}$ is a 10 dimensional vector.
}%
{%
  Musashi実機の左腕を用いて, 筋長制御・状態推定・オンライン学習・異常検知・筋破断確認について実験を行う.
  本節では, $\bm{\theta}$は5自由度, $\{\bm{l}, \bm{f}\}$は10自由度である.
}%

\begin{figure}[t]
  \centering
  \includegraphics[width=0.95\columnwidth]{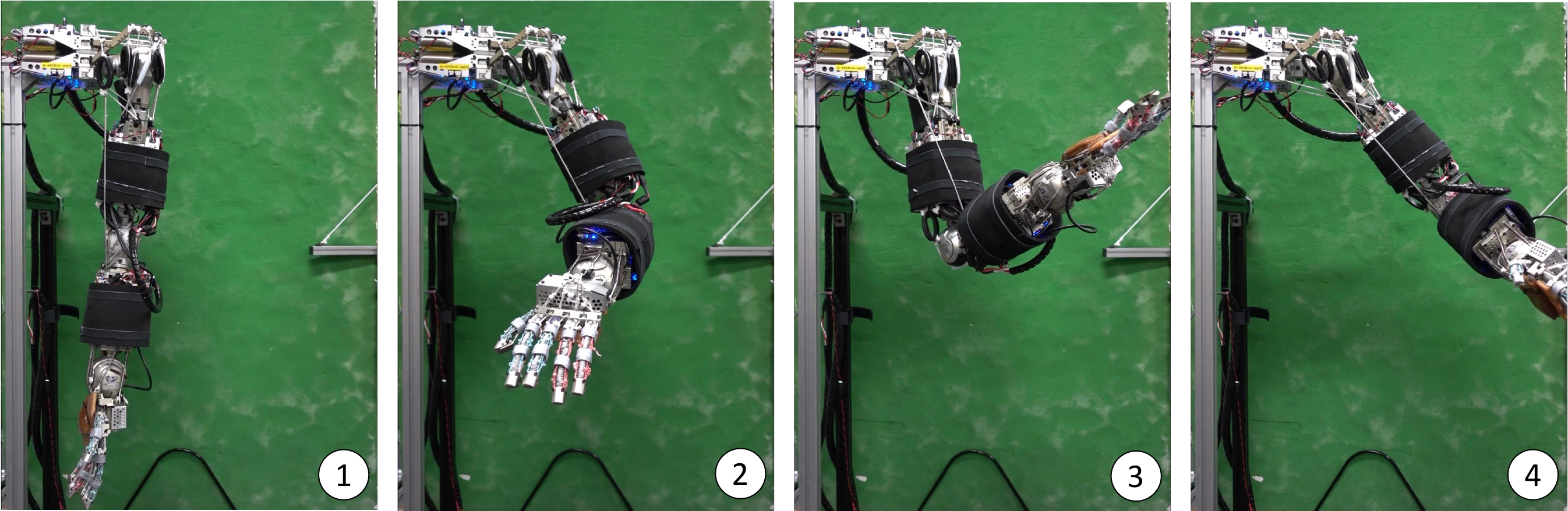}
  \vspace{-1.0mm}
  \caption{Actual robot experiment of online learning: these pictures show the random movements after online learning of RMAE without any muscle ruptured.}
  \label{figure:actual-state-appearance}
  \vspace{-1.0mm}
\end{figure}

\begin{figure}[t]
  \centering
  \includegraphics[width=0.9\columnwidth]{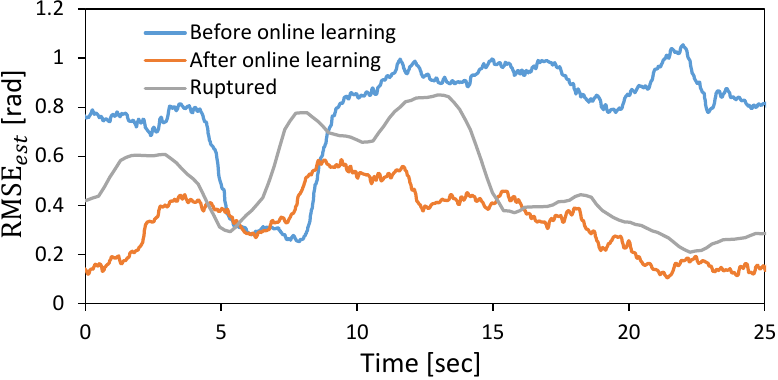}
  \vspace{-1.0mm}
  \caption{Actual robot experiment of state estimation: comparison of transitions of $\textrm{RMSE}_{est}$ among before online learning, after online learning, and when muscle \#9 is ruptured.}
  \label{figure:actual-state-compare}
  \vspace{-1.0mm}
\end{figure}

\begin{figure}[t]
  \centering
  \includegraphics[width=0.9\columnwidth]{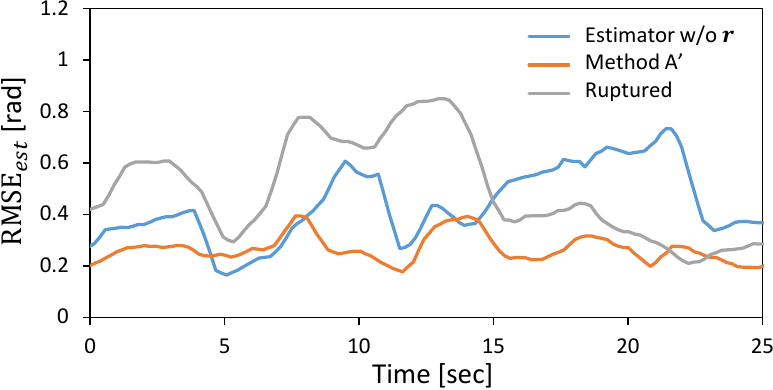}
  \vspace{-1.0mm}
  \caption{Actual robot experiment of state estimation with muscle \#9 ruptured: comparison of transitions of $\textrm{RMSE}_{est}$ between after online learning with and without $\bm{r}$.}
  \label{figure:actual-state-learning2}
  \vspace{-1.0mm}
\end{figure}


\subsubsection{Online Learning and State Estimator Experiment} \label{subsubsec:actual-state-estimation}
\switchlanguage%
{%
  We show the transitions of state estimation errors $\textrm{RMSE}_{est}$ before and after online learning of RMAE with the actual robot of Musashi (\figref{figure:actual-state-appearance}) during the evaluation experiment as in \secref{subsubsec:sim-state-estimation}, in \figref{figure:actual-state-compare}.
  \figref{figure:actual-state-compare} also shows the transition when muscle \#9 is ruptured (the current of the motor is set to zero) and when using method A' with RMAE after the online learning for the state estimation.
  The following evaluation experiment is basically the same as in \secref{subsubsec:sim-state-estimation}, but the number of random joint angles to be sent is increased to 7 and they are sent over 3 seconds.
  From the results, we can see that the error of the state estimation was much smaller after the online learning than before.
  When the muscle was ruptured, the error became slightly larger than after the online learning.
  The results of the state estimator in the actual robot including the subsequent experiments are shown in \tabref{table:actual-state-compare}.
  The average of $\textrm{RMSE}_{est}$ during 25 seconds of this motion, $\textrm{RMSE}^{ave}_{est}$, was 0.78 rad before online learning, 0.33 rad after online learning, and 0.50 rad after the muscle rupture.

  Next, \figref{figure:actual-state-learning2} shows the transition of $\textrm{RMSE}_{est}$ after online learning with the muscle ruptured.
  Here, we compare the transitions of the error when using method A' in \secref{subsec:state-estimation} (the method with the best performance in \secref{subsubsec:sim-state-estimation}) after the online learning in \secref{subsec:online-learning} with and without $\bm{r}$.
  We can see that the error was smaller with $\bm{r}$ than without $\bm{r}$.
  Without $\bm{r}$, the error was sometimes smaller or larger than before the online learning, and the error did not vary so much.
  $\textrm{RMSE}^{ave}_{est}$ was 0.44 rad without $\bm{r}$, and 0.27 rad with $\bm{r}$.
  Therefore, the online learning with $\bm{r}$ can maintain the same or lower error after the muscle rupture than that before the muscle rupture.
  %
}%
{%
  \secref{subsubsec:sim-state-estimation}と同様のオンライン学習を行い(\figref{figure:actual-state-appearance}), このオンライン学習の前と後のRMAEを使った際の, 実機における状態推定の誤差$\textrm{RMSE}_{est}$の遷移を\figref{figure:actual-state-compare}に示す.
  また, オンライン学習後のRMAEについて, 筋\#9が切れた状態について, 手法A'を使った場合についても, $\textrm{RMSE}_{est}$を\figref{figure:actual-state-compare}に示す.
  なお, 以降の評価時の動作については\secref{subsubsec:sim-state-estimation}と基本的には同様であるが, 送るランダムな関節角度数を7, 送る秒数を3秒に増やしている.
  グラフから, 学習後は学習前に比べて関節角度推定値の誤差が大きく減っていることがわかる.
  また, 筋が切れた際には, 学習後よりも少し誤差が増えてしまっていることがわかる.
  以降の実験を含めた実機における状態推定の結果を\tabref{table:actual-state-compare}に示す.
  25秒間の本動作における$\textrm{RMSE}_{est}$の平均$\textrm{RMSE}^{ave}_{est}$は, 学習前は0.78 rad, 学習後は0.33 rad, 筋破断時は0.50 radであった.

  次, その筋が切れた状態でオンライン学習を行った後に, 先と同じ実験を行ったときの$\textrm{RMSE}_{est}$の遷移を\figref{figure:actual-state-learning2}に示す.
  ここで, \secref{subsec:online-learning}における, 筋の破断情報$\bm{r}$が無かった場合とあった場合のオンライン学習後について, \secref{subsubsec:sim-state-estimation}で最も精度の高かった手法A'を用いた時の比較を行う.
  $\bm{r}$の情報があった場合は, 無かった場合に比べて誤差が少ないことがわかる.
  $\bm{r}$の情報がなかった場合は, 学習前より誤差が少ない場合や誤差が大きい場合があり, 大きくは変わらない.
  $\textrm{RMSE}^{ave}_{est}$は, $\bm{r}$の情報がない場合は0.44 rad, $\bm{r}$の情報がある場合は0.27 radであった.
  よって, $\bm{r}$の情報を含めてオンライン学習を継続した結果, 筋破断後も筋破断前と同程度かそれ以下の誤差を保つことができる.
  %
}%

\begin{table}[htb]
  \centering
  \caption{Results of evaluation experiments for state estimator in the actual robot: the averages of $\textrm{RMSE}_{est}$ during 25 seconds of the evaluation experiment,  $\textrm{RMSE}^{ave}_{est}$.}
  \vspace{0.1in}
  \scalebox{0.7}{
    \begin{tabular}{|l|c|c|c|c|c|} \hline
      Muscle state & \multicolumn{2}{c|}{w/o rupture} & \multicolumn{3}{c|}{w/ \#9 ruptured} \\ \hline
      Online learning & Before & After & Before & After, w/o $\bm{r}$ & After, w/ $\bm{r}$ \\ \hline
      Estimator w/o $\bm{r}$ [rad] & 0.78 & \textbf{0.33} & - & - & - \\
      Method A' [rad] & - & - & 0.50 & 0.44 & \textbf{0.27} \\\hline
    \end{tabular}
  }
  \label{table:actual-state-compare}
\end{table}

\begin{figure}[t]
  \centering
  \includegraphics[width=0.9\columnwidth]{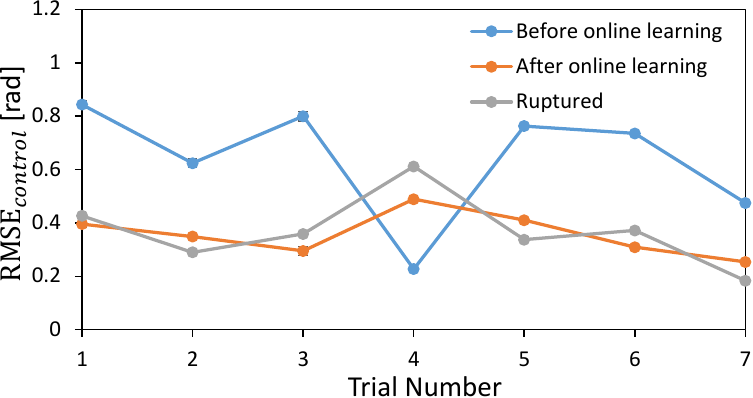}
  \vspace{-1.0mm}
  \caption{Actual robot experiment of controller: comparison of transitions of $\textrm{RMSE}_{control}$ among before online learning, after online learning, and when muscle \#9 is ruptured.}
  \label{figure:actual-control-compare}
  \vspace{-1.0mm}
\end{figure}

\begin{figure}[t]
  \centering
  \includegraphics[width=0.9\columnwidth]{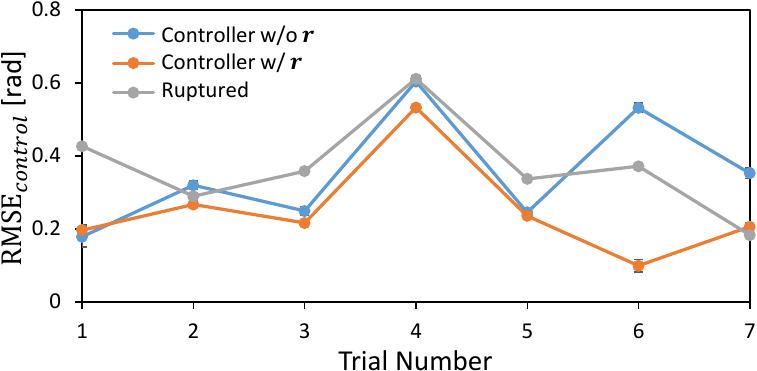}
  \vspace{-1.0mm}
  \caption{Actual robot experiment of controller with muscle \#9 ruptured: comparison of transitions of $\textrm{RMSE}_{control}$ between after online learning with and without $\bm{r}$.}
  \label{figure:actual-control-learning2}
  \vspace{-1.0mm}
\end{figure}

\subsubsection{Online Learning and Controller Experiment} \label{subsubsec:actual-control}
\switchlanguage%
{%
  We execute the evaluation experiment as in \secref{subsubsec:actual-state-estimation} to see the control error.
  Here, RMAEs before and after online learning in \secref{subsubsec:actual-state-estimation} are used for the controller, and $\textrm{RMSE}_{control}$ is calculated after each motion at rest.
  In this section, we perform the same motion five times, and display the average and variance of the errors.
  The results are shown in \figref{figure:actual-control-compare}.
  Regarding RMAE after the online learning, we also show the results of the controller in \secref{subsec:controller} with $\bm{r}$, where muscle \#9 is ruptured.
  Except for the fourth joint angle, we can see that the error was smaller after the online learning than before.
  In addition, the error did not change significantly when the muscle is ruptured, unlike the state estimation.
  This seems to be due to the fact that the online learning was successful for the robot state without muscle \#9.
  Also, the variance of the error was very small.
  The results of the controller in the actual robot including the subsequent experiments are shown in \tabref{table:actual-control-compare}.
  The average of $\textrm{RMSE}_{control}$ for the seven joint angles, $\textrm{RMSE}^{ave}_{control}$, was 0.64 rad before online learning, 0.36 rad after online learning, and 0.37 rad after the muscle rupture.

  Next, \figref{figure:actual-control-learning2} shows the average of $\textrm{RMSE}_{control}$ after the online learning with the muscle ruptured.
  Here, we compare the online learnings in \secref{subsec:online-learning} with and without $\bm{r}$.
  Note that the muscle length is controlled by including $\bm{r}$ in \secref{subsec:controller}.
  Although the results differed depending on the posture, we see that the error was smaller with $\bm{r}$ than without, and that the results were equivalent to those of the simulation.
  $\textrm{RMSE}^{ave}_{control}$ was 0.35 rad without $\bm{r}$ and 0.25 rad with $\bm{r}$.
}%
{%
  前節と同様にランダムな7つの関節角度$\bm{\theta}^{rand}$を$\bm{\theta}^{ref}$に指定し, これを順に3秒で送る動作を行う.
  このとき, 前節のオンライン学習前と学習後のRMAEを使って\secref{subsec:controller}を実行し, それぞれの動作後の静止時における$\textrm{RMSE}_{control}$を計算する.
  本節では, 同じ動作を5回行い, その平均と分散を計算し表示し, その結果を\figref{figure:actual-control-compare}に示す.
  また, オンライン学習後のRMAEについて, 筋\#9が切れた状態について, $\bm{r}$の情報がある状態で\secref{subsec:controller}を実行した場合についても結果を示す.
  4つ目の関節角度を除き, 基本的に学習後は学習前よりも誤差が小さくなっていることがわかる.
  また, 筋が切れた際は状態推定とは異なりあまり大きく誤差に変化はなかった.
  これは, 筋\#9を使わない状態についても学習が上手くいっているためだと考えられる.
  また, 誤差の分散は非常に小さい.
  以降の実験を含めた実機における制御の結果を\tabref{table:actual-control-compare}に示す.
  7回の本動作における$\textrm{RMSE}_{control}$の平均$\textrm{RMSE}^{ave}_{control}$は, 学習前は0.64 rad, 学習後は0.36 rad, 筋破断時は0.37 radであった.

  次に, その筋が切れた状態でオンライン学習を行った後に, 先と同じ実験を行ったときの$\textrm{RMSE}_{control}$を\figref{figure:actual-control-learning2}に示す.
  ここで, \secref{subsec:online-learning}における, 筋の破断情報$\bm{r}$があった場合となかった場合のオンライン学習について比較を行う.
  なお, \secref{subsec:controller}には破断情報$\bm{r}$を含めて筋長制御を行う.
  姿勢によって大きく異なるが, $\bm{r}$の情報がある場合はない場合よりも誤差が小さいことがわかり, シミュレーションと同等の結果であることがわかる.
  $\textrm{RMSE}^{ave}_{control}$は, $\bm{r}$がない場合は0.35 rad, $\bm{r}$がある場合は0.25 radであった.
}%

\begin{table}[htb]
  \centering
  \caption{Results of evaluation experiments for controller in the actual robot: the averages of $\textrm{RMSE}_{control}$ for the seven joint angles of the evaluation experiment.}
  \vspace{0.1in}
  \scalebox{0.7}{
    \begin{tabular}{|l|c|c|c|c|c|} \hline
      Muscle state & \multicolumn{2}{c|}{w/o rupture} & \multicolumn{3}{c|}{w/ \#2 ruptured} \\ \hline
      Online learning & Before & After & Before & After, w/o $\bm{r}$ & After, w/ $\bm{r}$ \\ \hline
      Controller w/o $\bm{r}$ [rad] & 0.64 & \textbf{0.36} & - & - & - \\
      Controller w/ $\bm{r}$ [rad] & - & - & 0.37 & 0.35 & \textbf{0.25} \\\hline
    \end{tabular}
  }
  \label{table:actual-control-compare}
\end{table}

\begin{figure}[t]
  \centering
  \includegraphics[width=0.99\columnwidth]{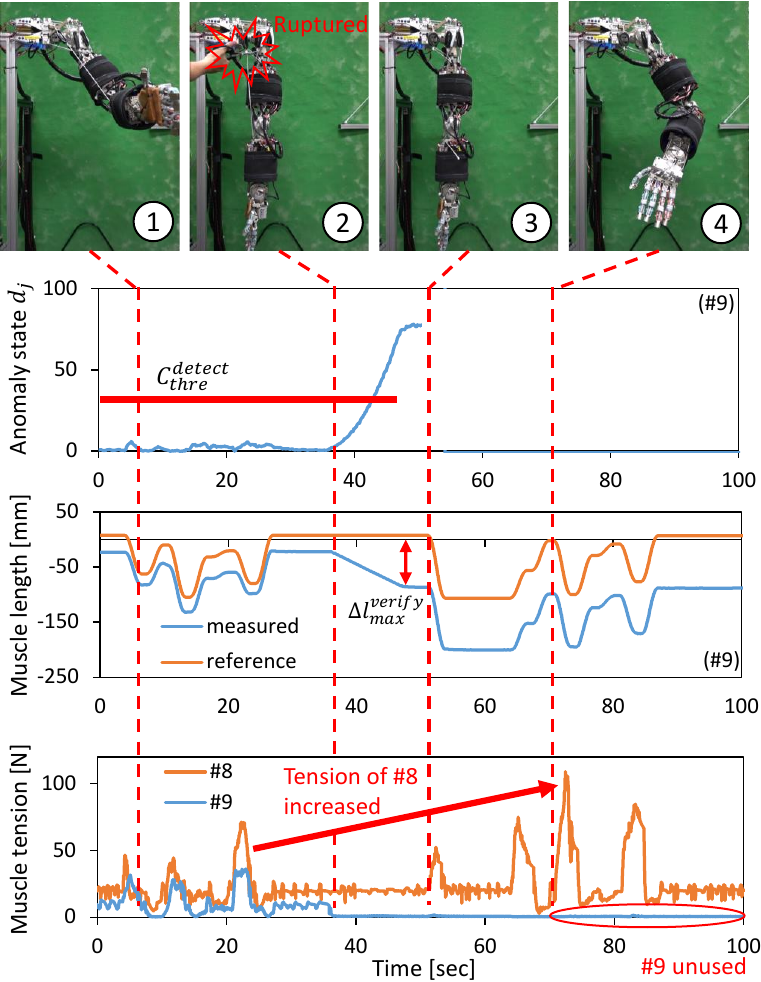}
  \vspace{-1.0mm}
  \caption{Actual robot experiment of anomaly detection and muscle rupture verification when the muscle wire of muscle \#9 is ruptured: the transitions of anomaly state $d_j$ and muscle length of muscle \#9, and transition of muscle tension of muscle \#8 and \#9.}
  \label{figure:actual-anomaly1}
  \vspace{-1.0mm}
\end{figure}

\begin{figure}[t]
  \centering
  \includegraphics[width=0.99\columnwidth]{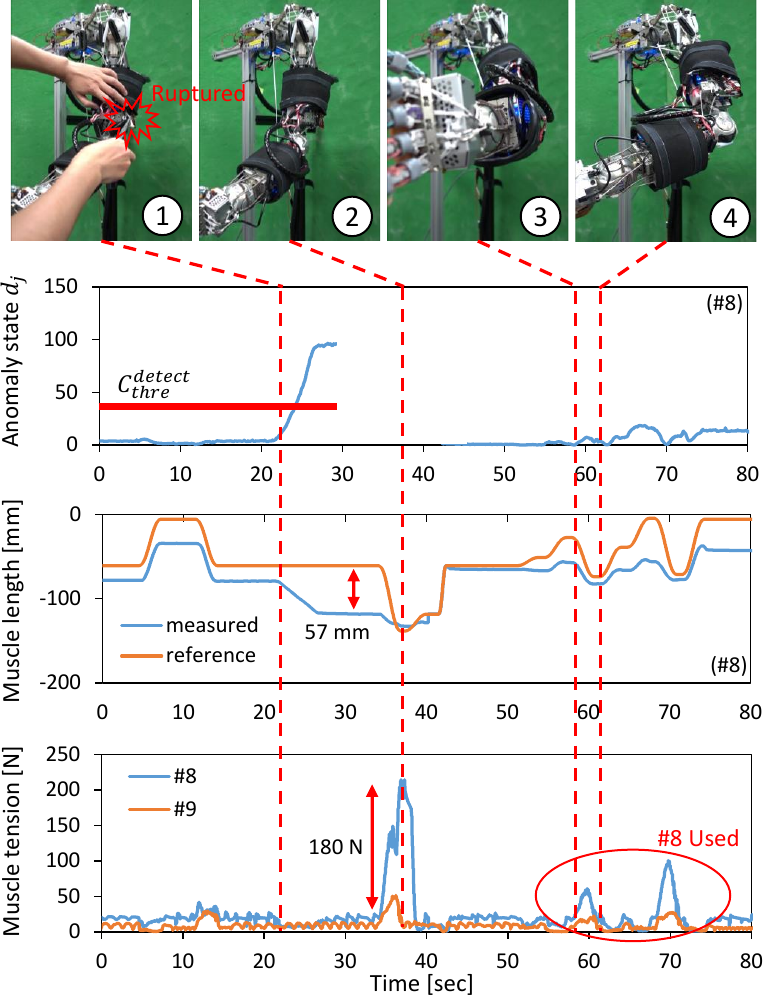}
  \vspace{-1.0mm}
  \caption{Actual robot experiment of anomaly detection and muscle rupture verification when the endpoint of muscle \#8 is ruptured: the transitions of anomaly state $d_j$ and muscle length of muscle \#8, and transition of muscle tension of muscle \#8 and \#9.}
  \label{figure:actual-anomaly2}
  \vspace{-1.0mm}
\end{figure}

\begin{figure*}[t]
  \centering
  \includegraphics[width=1.99\columnwidth]{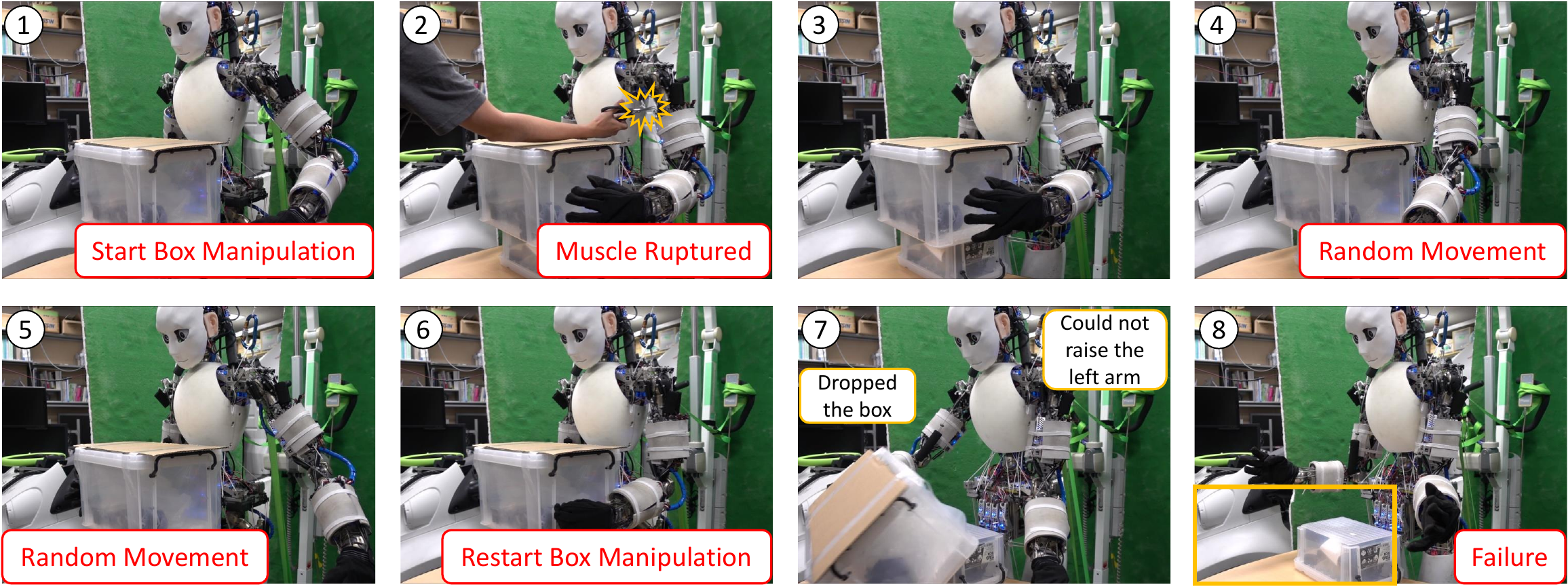}
  \vspace{-1.0mm}
  \caption{Continuous motion experiment by Musashi without muscle rupture verification: Musashi starts box manipulation, muscle rupture occurs, RMAE is updated online during random movements, Musashi restarts box manipulation, but Musashi dropped the box because its left arm could not be raised as expected.}
  \label{figure:integrated-sequence-without}
  \vspace{-1.0mm}
\end{figure*}

\begin{figure*}[t]
  \centering
  \includegraphics[width=1.99\columnwidth]{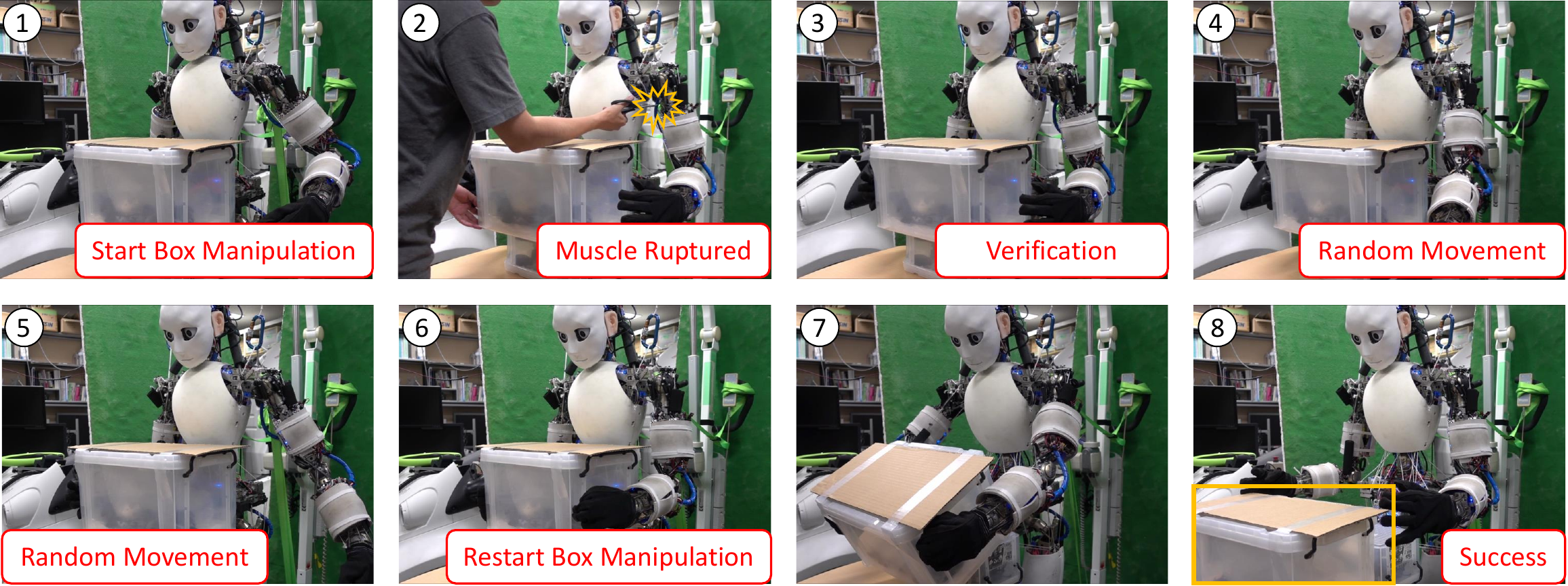}
  \vspace{-1.0mm}
  \caption{Continuous motion experiment by Musashi with muscle rupture verification: Musashi starts box manipulation, muscle rupture occurs, muscle rupture verification is executed to notify muscle rupture state $\bm{r}$ to all components, RMAE is updated online during random movements, Musashi restarts box manipulation, and Musashi succeeded in putting the box down with both arms.}
  \label{figure:integrated-sequence-with}
  \vspace{-1.0mm}
\end{figure*}

\subsubsection{Anomaly Detection and Muscle Rupture Verification Experiment} \label{subsubsec:actual-rupture}
\switchlanguage%
{%
  In this experiment, a muscle wire is actually cut and we test for anomaly detection and muscle rupture verification methods.

  First, we investigate how the anomaly detection and muscle rupture verification work when muscle \#9 is cut with scissors.
  The degree of anomaly $d_j$ and muscle length of muscle \#9, and muscle tensions of muscles \#8 and \#9 are shown in \figref{figure:actual-anomaly1} for the following series of motions: a random motion sending random joint angles, cutting the muscle, and then the random motion again, in order.
  Note that muscles \#8 and \#9 are synergistic muscles with similar moment arms.
  During the random motion, the muscle tension of \#8 and \#9 was about 69 N and 38 N at most.
  When muscle \#9 was cut, the muscle tension became zero, the muscle length was gradually shortened due to $\bm{f}^{bias}$, and $d_j$ exceeded $C^{detect}_{thre}=30.0$ ($d_j$ of other muscles did not change much).
  The difference between the target and measured muscle lengths stopped at $\Delta{l}^{verify}_{max}=100$ [mm], which was set as the maximum value.
  After that, the target muscle length changed to $l^{cur}_{i}-\Delta{l}^{verify}_{pull}$ ($\Delta{l}^{verify}_{pull}=10$ [mm]), but the muscle tension did not change at all because the muscle was broken.
  Therefore, \#9 was judged to be ruptured (state (a)) and $\bm{r}$ was notified to all components.
  After that, the robot could move its joints as usual, but no tension was applied to muscle \#9 and about 100 N was applied to muscle \#8 at most.

  We next investigate how the anomaly detection and muscle rupture verification work when the end point of muscle \#8 has been taken off.
  The degree of anomaly $d_j$ and muscle length of muscle \#8, and muscle tensions of muscles \#8 and \#9 are shown in \figref{figure:actual-anomaly2} for a series of motions of bending the elbow, cutting the muscle, and then performing a random motion, in order.
  After the muscle rupture of \#8, the muscle length was gradually shortened by $f_{bias}$, and $d_j$ of muscle \#8 exceeded $C^{detect}_{thre}=30.0$ ($d_j$ of other muscles did not change much).
  The difference between the target and measured muscle lengths stopped at 57 mm, which was smaller than the maximum value $\Delta{l}^{verify}_{max}=100$ [mm].
  Thereafter, the target muscle length was changed to $l^{cur}_{i}-\Delta{l}^{verify}_{pull}$ ($\Delta{l}^{verify}_{pull}=10$ [mm]), and $l^{cur}_{i}$ followed it, resulting in an increase in muscle tension of about 180 N.
  Since $\Delta{f}_{i}>\Delta{f}^{verify}_{thre}$ ($\Delta{f}^{verify}_{thre}=10$ [N]) and $\textrm{abs}(l^{cur}_{i}-l^{ref}_{i})>\Delta{l}^{verify}_{thre}$ ($\Delta{l}^{verify}_{thre}=30$ [mm]), muscle \#8 was determined to be in state (b) and the muscle length was initialized.
  After that, the robot could move its joints as usual, and a muscle tension of about 120 N was applied to muscle \#8.
}%
{%
  実機の筋を実際に切り, 異常検知と筋破断確認について実験を行う.

  まず, 筋\#9をハサミで切ったときに, どのように異常検知・筋破断確認が働くかを調べる.
  通常動作を行い, その後筋を切り, また通常動作を行うという一連の動きをしたときの, 筋\#9の異常度$d_j$と筋長, 筋\#8と\#9の筋張力を\figref{figure:actual-anomaly1}に示す.
  なお, \#8と\#9は互いに似通ったモーメントアームを持つ共同筋である.
  通常動作時には\#8の筋張力が最大で約70 N, \#9の筋張力が最大で約35 N程度まで出ている.
  \#9を切ると, 筋張力0になり, 筋長が$\bm{f}^{bias}$によって徐々に短くなり, 異常度は上がり, 途中で$C^{detect}_{thre}$を超えた(なお, 他の筋の異常度はほとんど変化しなかった).
  指令筋長と測定された筋長の差は最大値として設定した$\Delta{l}^{verify}_{max}$のところで止まっている.
  その後, 筋長指令が$l^{cur}_{i}-\Delta{l}^{verify}_{pull}$の位置まで変化しするが, 筋が切れているため筋張力は一切変化しない.
  そのため\#9は破断した(状態(a))と判断され, $\bm{r}$が全コンポーネントに通達された.
  その後通常の動作を行うと, ロボットは通常通り関節を動かせるものの, \#9には一切の筋張力はかからず, \#8は最大で約100 Nの力がかかっていた.

  次に, 筋\#8の末端を外したときに, どのように異常検知・筋破断確認が働くかを調べる.
  肘を曲げ, その状態で筋を切り, その後通常の動作を行うという一連の動きをしたときの, 筋\#8の異常度$d_j$と筋長, 筋\#8と\#9の筋張力を\figref{figure:actual-anomaly2}に示す.
  \#8の筋破断後, 筋長が$\bm{f}^{bias}$によって徐々に短くなり, 異常度は上がり, 途中で$C^{detect}_{thre}$を超えた(なお, 他の筋の異常度はほとんど変化しなかった).
  指令筋長と測定された筋長の差は最大値として設定した100 mmよりも短い57 mmのところで止まっている.
  その後, 筋長指令が$l^{cur}_{i}-\Delta{l}^{verify}_{pull}$の位置まで変化し, $l^{cur}_{i}$はそれに追従し, 筋張力が180 N程度増大した.
  $\Delta{f}_{i}>\Delta{f}^{verify}_{thre}$かつ$\textrm{abs}(l^{cur}_{i}-l^{ref}_{i})>\Delta{l}^{verify}_{thre}$であるため, \#8は状態(b)と判断され, 筋長の初期化が行われた.
  その後通常の動作を行うと, ロボットは通常通り関節を動かすことができ, \#9だけでなく, \#8にも120 N程度の力がかかっていることがわかる.
}%

\subsection{Continuous Motion Experiment by Musashi}
\switchlanguage%
{%
  The musculoskeletal humanoid Musashi is used to perform a box manipulation experiment, and a series of events including muscle rupture, anomaly detection, muscle rupture verification, and online learning occur.
  The robot holds a large box with both hands and puts it down.
  During this experiment, online learning is executed at all times.

  Muscle \#9 of the left arm is cut with scissors before holding the box.
  Here, we compare the cases of whether the muscle rupture verification in \secref{subsubsec:actual-rupture} is executed or not.
  With the muscle rupture verification, $\bm{r}$ is notified to all the components, while without it, $\bm{r}$ is not taken into account for online learning, controller, and state estimation even if the muscle is ruptured.
  Then, as in \secref{subsubsec:actual-control}, the left arm is moved by random joint angles and muscle tensions for online learning of RMAE.
  We then resume the box manipulation experiment (online learning continues to work during this time).
  Here, we set $N^{online}_{thre}=2$ so that the online learning starts immediately.

  \figref{figure:integrated-sequence-without} shows the case without the muscle rupture verification and \figref{figure:integrated-sequence-with} shows the case with the muscle rupture verification.
  When the muscle rupture verification was not performed, online learning proceeded with muscle length of one muscle with a strange value, so the robot could not raise its left hand as expected and dropped the box.
  In contrast, in the case with the muscle rupture verification, both hands worked well together and the box was manipulated correctly after resumption of the manipulation, because the muscle rupture information $\bm{r}$ was notified and the online learning progressed correctly.
}%
{%
  筋骨格ヒューマノイドMusashiを用いて箱のマニピュレーション実験を行い, その際に筋破断・異常検知・筋破断確認・オンライン学習という一連の流れを行った.
  大きな箱を両手で持ち, 移動させる動作を行う.
  これまで開発したコンポーネントを全て実行しておき, 箱を持つ手前で左腕の筋\#9をハサミで切断する.
  このとき, \secref{subsubsec:actual-rupture}の筋破断確認を行う場合と行わない場合で比較する.
  行った場合は$\bm{r}$が全コンポーネントに通達されるのに対して, 行わない場合は筋が切れても$\bm{r}$を考慮することはない.
  その後, \secref{subsubsec:actual-control}と同様にランダムな関節角度・筋張力によって左腕を動かすことで, RMAEがオンライン学習されていく.
  その後元通りに実験を再開する.
  このとき, すぐにオンライン学習が再開されるよう, $N^{online}_{thre}=2$とした.

  その全体の動作シーケンスについて, 筋破断確認を行わない場合を\figref{figure:integrated-sequence-without}に, 行う場合を\figref{figure:integrated-sequence-with}に示す.
  筋破断確認を行わない場合は, 一つの筋長がおかしな値のまま学習が進むため, ランダム動作による学習後に動きを再開する過程で, 思ったように手が上がらなくなり, 箱を落としてしまっていることがわかる.
  これに対して筋破断確認を行う場合は, 筋破断情報$\bm{r}$が通達され, 正しく学習が進むため, 動作の再開後, 右手と左手が正しく協調して動き, 箱を正しくマニピュレーションできていることがわかる.
}%

\section{Discussion} \label{sec:discussion}
\switchlanguage%
{%
  Based on the experimental results of this study, we discuss the effects of muscle rupture on online learning, state estimation, and control using RMAE.

  First, it is possible to see the ideal performance of RMAE in the simple 1 DOF simulation experiments.
  In the online learning method, RMAE can reduce the error of the joint angle estimation by one third in 4 minutes.
  In the state estimation, the error can be greatly reduced by using the muscle rupture information $\bm{r}$ when the muscle breaks.
  Although the computational cost of method A' is more expensive, it can achieve higher accuracy than method A.
  It is also found that while the estimation error is extremely large when updating RMAE online without $\bm{r}$ after a muscle rupture, the estimation error can be reduced even further by changing the online learning method using $\bm{r}$.
  This tendency is also true for the controller using RMAE, where online learning of RMAE significantly reduces the control error.
  By using the information of $\bm{r}$, the control error can be further reduced by online learning after the muscle rupture.

  Next, in the experiment on the 5 DOFs of the left shoulder and elbow of Musashi, we found out by how much the performance of components using RMAE is affected by the characteristics of the actual robot.
  For the state estimation, we can see that the tendency of the estimation error is almost the same as that of the simulation, and that the online learning of RMAE improves the error significantly.
  However, although the peak of the error can be suppressed, we cannot see as large a difference between methods A and A' as in the simulation.
  This is considered to be because RMAE is not updated as completely as in the simulation, and therefore the difference between methods A and A' is lost in the error of RMAE.
  In addition, the control error can be reduced by online learning, and online learning with $\bm{r}$ is effective as well.
  However, in general, the difference between with and without $\bm{r}$ is smaller than that of the simulation, and the error depends greatly on the posture of the robot.
  It is also found that the anomaly detection and muscle rupture verification are able to correctly determine the current state of the muscles, to recognize muscles that are not usable and notify $\bm{r}$ to all components, and to initialize and reuse muscles that are usable but are offset in the initial muscle length.

  Finally, in the box manipulation experiment describing the overall flow of this study, it is found that there is a significant difference in task execution depending on the presence or absence of the muscle rupture information $\bm{r}$.
  If the online learning process is continued without $\bm{r}$, the abnormal sensor values of the ruptured muscle deteriorate the entire network, and the target task fails.

  From this study, we found that in order to take advantage of the redundant muscle arrangement, it is necessary to recognize the muscle rupture and process it appropriately to perform online learning, state estimation, and control of the flexible body structure.
  It is also important to construct a whole system that includes replaceable hardware, anomaly detection, and muscle rupture verification.
  We believe that our method may be one of the methods that can cope with malfunctions of the robot, such as the failure of some sensors and drive systems.
  In addition, because there are some muscles that cannot be compensated if they are ruptured, our research limits the muscles that can be ruptured (\#2, \#8, and \#9).
  In future works, we would like to develop a design optimization method of the muscle arrangement so that any muscle rupture can be compensated.
}%
{%
  本研究の実験結果から, RMAEのオンライン学習・状態推定・制御について, 筋が破断した際におけるそれらコンポーネントへの影響について考察する.

  まず, 1自由度のシミュレーション実験では, 理想的にどのようなパフォーマンスの違いがあるのかを知ることが可能である.
  オンライン学習ではRMAEをオンラインで学習させることで, 4分間で関節角度推定の誤差が1/3程度まで減少した.
  状態推定においては, 筋が切れた際は, $\bm{r}$の情報を用いることで大幅に誤差を軽減できる.
  その中でも, 繰り返し計算を行う手法A'は計算コストは高いものの, 手法Aよりも高い精度を出すことが可能である.
  また, 筋破断後, $\bm{r}$を用いずにオンライン学習させると, 推定誤差が非常に大きくなるのに対して, $\bm{r}$を利用して学習方法を変化させることで, 筋破断前よりもさらに推定誤差を減らすことが可能なことがわかった.
  この傾向はRMAEを使った制御についても同様であり, RMAEのオンライン学習により大幅に制御誤差を減らすことができ, $\bm{r}$の情報を用いることで筋破断後のオンライン学習によってもさらに誤差を減らすことが可能である.

  次に, 5自由度のMusashiの左肩・肘における実験では, 実機に置いてその性能がどの程度変化するかを知ることが可能である.
  状態推定についてはシミュレーションとほぼ同じ傾向にあり, RMAEの学習によって大幅に誤差を改善することができ, $\bm{r}$の情報を含めたオンライン学習が非常に有効であった.
  しかし, 誤差のピークは抑えられるものの, 手法AとA'についてはシミュレーションほど大きな差を見ることは出来なかった.
  これは, シミュレーションほど綺麗にRMAEが学習されるわけではないため, その誤差に手法AとA'の違いが紛れてしまったと考えられる.
  また, 制御についてもオンライン学習により誤差を低減することができ, 同様に$\bm{r}$の情報を含めたオンライン学習は有効であった.
  しかし, 全体的に$\bm{r}$を含めた場合は含めない場合における誤差の変化はシミュレーションよりも小さく, 自由度が多いため誤差は姿勢に大きく依存していた.
  また, 異常検知・筋破断確認は正確に現在の筋状態を判断することができ, 使えない筋を認識して$\bm{r}$を通達する, 使えるが初期筋長がオフセットしてしまった筋を初期化して再度利用することができることがわかった.

  最後に, 全体の流れを記述したbox manipulation実験では, 筋破断状態$\bm{r}$の有無によってタスク遂行に大きな差が出ることがわかった.
  $\bm{r}$がないままオンライン学習を実行し続けると, 破断した筋の異常なセンサ値がネットワーク全体に悪さをし, ネットワーク構造が崩れ, タスクが失敗してしまうことがわかった.

  本研究から, 冗長な筋配置の性質を最大限に利用するためには, 筋破断に気づき, それを適切に処理して, その柔軟な身体構造に関するオンライン学習・状態推定・制御等を行っていく必要があることがわかった.
  また, 交換可能なハードウェア・異常検知や筋破断確認等の全体システムも重要である.
  本手法の考え方は, 今後一部のセンサや駆動系が壊れた等のロボットの異常に対する対処法を与える一つの手段になると考えている.
  また, 現状一度切れたら関節を動かせなくなってしまう筋もあるため本研究では切る筋を限定していた.
  今後, どの筋が切れても大丈夫なような設計の最適化等についても開発を進めたい.
}%

\section{Conclusion} \label{sec:conclusion}
\switchlanguage%
{%
  In this study, we proposed a new learning control method making use of the characteristic that the redundant musculoskeletal humanoid can keep moving its joints even when one of its redundant muscles is ruptured.
  The redundant musculoskeletal autoencoder (RMAE), which expresses redundant intersensory relationships among joint angle, muscle tension, and muscle length, is constructed and used for the state estimation and control of the flexible and difficult-to-modelize body by updating RMAE online.
  Anomaly is detected from the prediction error of RMAE, and muscle rupture is verified by moving the muscles one by one.
  When a muscle rupture occurs, the online learning, state estimation, and control algorithms are modified based on the muscle rupture information to adapt to the current robot condition without destroying the current network structure of RMAE.
  Experimental results show that the presence or absence of the muscle rupture information significantly affects the errors in control and state estimation, and the effectiveness of this study was confirmed through a box manipulation experiment.

  In future works, we will continue to run these systems for a long time in order to construct a more robust robot system.
}%
{%
  本研究では, 筋骨格ヒューマノイドの冗長な筋が一本切れても関節を動かし続けられるという性質を最大限に利用した, 学習型制御手法について提案した.
  関節角度・筋張力・筋長の冗長な相互関係を記述するredundant musculoskeletal autoencoder (RMAE)を構築し, これをオンライン学習させ, 柔軟でモデル化困難な身体の状態推定・筋長制御に利用する.
  RMAEの予測誤差から異常検知を行い, 筋を一本ずつ動かして筋破断確認を行うことで, 筋破断を検知する.
  筋破断時には, RMAEのオンライン学習・状態推定・筋長制御のアルゴリズムを筋破断情報を元に変更することで, 現状のネットワーク構造を壊さずに, 現在の身体状態に適応していくことが可能である.
  筋破断情報の有無が顕著に制御・状態推定の誤差に影響を及ぼすことを実験から示し, 一連の継続的動作実験により本研究の有効性を確認した.

  今後, これらのシステムを長期的に走らせていき, よりロバストなロボットシステムの構築を目指したい.
}%

\bibliography{main}

\end{document}